\documentclass[10pt,journal,cspaper,compsoc]{./IEEEtran}

\usepackage{amssymb}
\usepackage{amsmath}
\usepackage{graphicx}
\usepackage{subfigure}
\usepackage{amsthm}
\usepackage{color}
\usepackage{multirow,bigstrut}
\usepackage{caption}
\DeclareMathOperator*{\argmax}{argmax}
\DeclareMathOperator*{\argmin}{argmin}

\newtheorem{theorem}{Theorem}

\newtheorem{definition}{Definition}

\ifCLASSOPTIONcompsoc
\else
\fi
\ifCLASSINFOpdf
\else
\fi
\begin{document}

    %
    \title{Optimizing Evaluation Metrics for Multi-Task Learning via the Alternating Direction Method of Multipliers}
    %
    %
    %
    %

    \author{Ge-Yang Ke,
        Yan Pan,
        Jian Yin,
        Chang-Qin Huang
            \thanks{Ge-Yang Ke, Yan Pan, and Jian Yin are with the School of
            Data Science and Computer Science, Sun Yat-sen University, Guangzhou 510006, China. Corresponding author: Yan Pan (panyan5@mail.sysu.edu.cn)}

            \thanks{Chang-Qin Huang is with the School of Information Technology in Education, South China Normal University, Guangzhou, 510631, China}
            }

    \markboth{IEEE TRANSACTIONS ON CYBERNETICS,~Vol.~, No.~, 2017}%
    {Ke \MakeLowercase{\textit{et al.}}: Optimizing Evaluation Metrics for Multi-Task Learning via the Alternating Direction Method of Multipliers}
    %


    
    \IEEEcompsoctitleabstractindextext{%
        
        \begin{abstract}
Multi-task learning (MTL) aims to improve the generalization
performance of multiple tasks by exploiting the shared factors among
them. Various metrics (e.g., F-score, Area Under the ROC Curve) are
used to evaluate the performances of MTL methods. Most existing MTL
methods try to minimize either the misclassified errors for
classification or the mean squared errors for regression. In this
paper, we propose a method to directly optimize the evaluation
metrics for a large family of MTL problems. The formulation of MTL
that directly optimizes evaluation metrics is the combination of two
parts: (1) a regularizer defined on the weight matrix over all
tasks, in order to capture the relatedness of these tasks; (2) a sum
of multiple structured hinge losses, each corresponding to a surrogate
of some evaluation metric on one task. This formulation is
challenging in optimization because both of its parts are
non-smooth. To tackle this issue, we propose a novel optimization
procedure based on the alternating direction scheme of multipliers,
where we decompose the whole optimization problem into a sub-problem
corresponding to the regularizer and another sub-problem
corresponding to the structured hinge losses. For a large family of MTL problems, the first sub-problem has closed-form solutions. To solve the second sub-problem, we
propose an efficient primal-dual algorithm via coordinate ascent.
Extensive evaluation results demonstrate that, in a large family of
MTL problems, the proposed MTL method of directly optimization
evaluation metrics has superior performance gains against the
corresponding baseline methods.

        \end{abstract}

        }

        \maketitle

        \IEEEdisplaynotcompsoctitleabstractindextext

        %
        \IEEEpeerreviewmaketitle

\section{Introduction}
Recently, considerable research has been devoted to \textit{Multi-Task Learning (MTL)}, a problem of improving the
generalization performance of multiple tasks by utilizing the shared
information among them. MTL has been widely-used in various
applications, such as natural language processing~\cite{Ando},
handwritten character recognition~\cite{Obozinski,Quadrianto}, scene recognition~\cite{TCYB-1} and
 medical diagnosis~\cite{Bi}. Many MTL methods have been proposed in the literature~\cite{Caruana,Evgeniou05,Yu,Zhang,Gong,Kang,Liu,Obozinski,Zhou,Ando,Chen10,Chen11,Pan3,TCYB-1,TCYB-2,TCYB-3,TCYB-4,TCYB-MTL1,TCYB-MTL2}.

In this paper, we consider MTL for classification or regression problems. Note that either a multi-class classification problem or a multi-label learning problem can be regarded as an MTL problem\footnote{As an illustrative example, we consider a multi-label classification problem with instances $\{x_1,x_2,x_3,x_4,x_5\}$ that $x_1$ belongs to classes \textcolor[rgb]{1,0,0}{$a$} and \textcolor[rgb]{0,1,0}{$b$}, $x_2$ belongs to classes \textcolor[rgb]{0,1,0}{$b$} and \textcolor[rgb]{0,0,1}{$c$}, $x_3$ belongs to class \textcolor[rgb]{0,0,1}{$c$}, $x_4$ belongs to class \textcolor[rgb]{1,0,0}{$a$}, $x_5$ belongs to classes \textcolor[rgb]{1,0,0}{$a$}, \textcolor[rgb]{0,1,0}{$b$} and \textcolor[rgb]{0,0,1}{$c$}. This problem can be regarded as an MTL problem with three tasks, where the training sets for each of these tasks are:\\
\begin{displaymath}
\begin{split} (x_1,\textcolor[rgb]{1,0,0}{1}),(x_2,\textcolor[rgb]{1,0,0}{0}),(x_3,\textcolor[rgb]{1,0,0}{0})
    ,(x_4,\textcolor[rgb]{1,0,0}{1}),(x_5,\textcolor[rgb]{1,0,0}{1})\\
     (x_1,\textcolor[rgb]{0,1,0}{1}),(x_2,\textcolor[rgb]{0,1,0}{1}),(x_3,\textcolor[rgb]{0,1,0}{0})
    ,(x_4,\textcolor[rgb]{0,1,0}{0}),(x_5,\textcolor[rgb]{0,1,0}{1})\\ (x_1,\textcolor[rgb]{0,0,1}{0}),(x_2,\textcolor[rgb]{0,0,1}{1}),(x_3,\textcolor[rgb]{0,0,1}{1})
    ,(x_4,\textcolor[rgb]{0,0,1}{0}),(x_5,\textcolor[rgb]{0,0,1}{1})\\
\end{split}
\end{displaymath}
The first/second/third task is a binary classification problem of an instance belonging to class \textcolor[rgb]{1,0,0}{$a$}/ \textcolor[rgb]{0,1,0}{$b$}/\textcolor[rgb]{0,0,1}{$c$} or not. Hence, a multi-label learning problem is a special case of an MTL problem. Similarly, we can verify that a multi-class classification problem can also be regarded as an MTL problem.}. Most of the existing MTL methods focus on minimizing either a convex surrogate (e.g. the hinge loss or the logistic loss) of the $0$-$1$ errors for multi-task classification, or the mean squared errors for multi-task regression. On the other hand, in practice, several evaluation metrics other than misclassified errors or mean squared errors are used the evaluation of MTL methods, e.g., F-score, Precision, Recall, Area Under the ROC Curve (AUC), Mean Average Precision. For example, in the cases of MTL on imbalanced data (e.g., in a task, the number of negative samples is much larger than that of the positive samples), cost-sensitive MTL or MTL for ranking, these metrics are more effective in performance evaluation than the standard misclassified errors or the mean squared errors. However, due to the computational difficulties, few learning techniques have been developed to directly optimize these evaluation metrics in MTL.

In this paper, we propose an approach to directly optimizing the evaluation metrics in MTL, which can be applied to a large family of MTL problems. Specifically, for an MTL problem with $m$ tasks (the $i$th task is associated with a training set $\{(\bold{x}_j^{(i)},\bold{y}_j^{(i)})\}_{j=1}^{n_i}$, $i=1,2,...,m$, $n_i$ represents the number of training samples for the $i$th task), we consider a generic formulation in the following:
\begin{equation}\label{MTL}
\min_{\bold{W}} {\Omega}(\bold{W})+\lambda \sum_{i=1}^{m}\mathcal{L}(\bold{W}_{.i};\{(\bold{x}_j^{(i)},\bold{y}_j^{(i)})\}_{j=1}^{n_i}),
\end{equation}
where $\bold{W}$ is the weight matrix with $m$ columns $\bold{W}_{.1}$, $\bold{W}_{.2}$, ..., $\bold{W}_{.m}$, $\lambda>0$ is a trade-off parameter. This formulation is the linear combination of two
parts. The first part is a regularizer ${\Omega}(\bold{W})$ defined on the weight matrix $\bold{W}$ over all
tasks, in order to leverage the relatedness of these tasks. Examples of this kind of regularizer include the trace-norm, the $\ell_{1,1}$-norm or the $\ell_{2,1}$-norm on $\bold{W}$. The second part in the formulation is a sum of multiple loss functions, each corresponds to one task. In order to directly optimize a specific evaluation metric, we consider the hinge loss functions for structured outputs~\cite{Struct-SVM,SVM-Perf,SVM-MAP,TCYB-5,TCYB-6}, which are surrogates of a specific evaluation metric.

Such a formulation in (\ref{MTL}) includes a large family of MTL problems. Since the two parts in (\ref{MTL}) are usually non-smooth, the optimization problem (\ref{MTL}) is difficult to solve. To tackle this issue, we propose a novel optimization
procedure based on the alternating direction scheme of multipliers (ADMM~\cite{ADMM,ALM}), which is widely used in various machine learning problems (e.g.,~\cite{Pan,Pan2,Pan3,Xia}). We decompose the whole optimization problem in (\ref{MTL}) into two simpler sub-problems. The first sub-problem
corresponds to the regularizer. For commonly-used regularizers (e.g., the trace-norm, the $\ell_{2,1}$-norm) in MTL, this sub-problem can be solved by close-form solutions. The second sub-problem corresponds to the structured hinge losses. To solve the second sub-problem, we propose an efficient primal-dual algorithm via coordinate ascent.

We conduct extensive experiments to evaluate the performances of the proposed MTL method. Experimental results show that the proposed method that optimizes a specific evaluation metric outperforms the corresponding MTL classification or MTL regression baseline methods by a clear margin.

\section{Related Work}\label{Related Work}
MTL is a wide class of learning problems.
Roughly speaking, the existing MTL methods can be divided into three main
categories: parameters sharing, common features sharing, and
low-rank subspace sharing.

In the methods with parameter sharing, all tasks are assumed to
explicitly share some common parameters. Representative methods in
this category include shared weight vectors~\cite{Evgeniou05},
hidden units in neural network~\cite{Caruana}, and common prior in
Bayessian models~\cite{Yu,Zhang}.

In the methods with common features sharing, task relatedness is
modeled by enforcing all tasks to share a common set of
features~\cite{Argyriou,Liu,Kim and Xing,Obozinski,Gong,Kang,Zhou}.
Representative examples are the methods which constrain the model
parameters (i.e., a weight matrix) of all tasks to have certain
sparsity patterns, for example, cardinality
sparsity~\cite{Obozinski}, group sparsity~\cite{Liu,Gong}, or
clustered structure~\cite{Kang,Zhou}.

The methods in the third category assume that all tasks lie in a
shared low-rank subspace~\cite{Ando,Chen10,Chen11}. A common assumption in these category of methods is
that most of the tasks are relevant while (optionally) there may exist a small
number of irrelevant (outlier) tasks. These methods pursue a low-rank
weight matrix that captures the underlying shared factors among tasks. Trace-norm regularization is commonly-used in these methods to
encourage the low-rank structure on the model parameters.

Most of the existing MTL methods are focused on designing regularizers or parameter sharing patterns to utilize the intrinsic relationships among multiple related tasks. These MTL methods usually try to optimize the classification errors or the mean squared errors for regression. In practice, various other metrics (such as F-score and AUC) are used in the evaluation of MTL methods. However, little effort has been devoted to optimize these evaluation metrics in the context of MTL except for the work~\cite{NIPS11}, in which the author proposed a hierarchical MTL formulation for structured output prediction in sequence segmentation. Since the regularizer used in~\cite{NIPS11} is decomposable, the hierarchical MTL problem can be decomposed into multiple independent tasks, each is a structure output learning problem with a simple regularizer. In this paper, we seek to directly optimize commonly-used evaluation metrics for MTL with possibly indecomposable regularizer, resulting in a generic approach that can be applied to a large family of MTL problems. Our formulation can be regarded as MTL for structure output prediction with an indecomposable regularizer.

The proposed methods in this paper are also related to the multi-label algorithms. There are various multi-label algorithm proposed in the literature, e.g., the RAkEL method that uses random $k$-label sets~\cite{Tsoumakas}, the MLCSSP method that spans the original label space by subset of labels~\cite{James}, the AdaBoostMH method based on AdaBoost~\cite{schapire2000boostexter}, the HOMER method based on the hierarchy of multi-label learners~\cite{Vlahavas}, the binary relevance (BR)~\cite{LP-BR} method, the label power-set (LP)~\cite{LP-BR} method, and the ensembles of classifier chains (ECC)~\cite{read2011classifier} method.

The proposed approach in this paper is to optimize the evaluation metrics in MTL. We refer the readers to Section \ref{PS} for the detailed introduction to the evaluation metrics related to the proposed approach.

\section{Notations}\label{notations}
We first introduce the notations to be used throughout this paper. We use bold upper-case characters (e.g., $\bold{M}$, $\bold{X}$, $\bold{W}$) to represent matrices, and bold lower-case characters (e.g., $\bold{x}$, $\bold{y}$) to represent vectors, respectively. For a matrix $\mathbf{M} \in {\mathbb{R}^{d  \times m }}$, we denote $\mathbf{M}_{ij}$ as the the element at the cross of the $i$th row and $j$th column in $\mathbf{M}$.
We denote $\mathbf{M}_{i \cdot}\in \mathbb{R}^{1  \times m }$ as the $i$th row of $\mathbf{M}$, and $\mathbf{M}_{\cdot j}\in \mathbb{R}^{d  \times 1 }$ as the $j$-th column of $\mathbf{M}$, respectively.

 We denote $||\mathbf{M}||_F$ as the Frobenius norm of $\mathbf{M}$ that $\|\mathbf{M}\|_F=\sqrt{\sum_{i=1}^d\sum_{j=1}^m (\mathbf{M}_{ij})^2}$.
Let $\|\bold{M}\|_{1,1} = \sum_{i=1}^{d}\sum_{j=1}^{m}
|\bold{M}_{ij}|$ be the $\ell_{1,1}$-norm of $\bold{M}$, where
$|\bold{M}_{ij}|$ is the absolute value of $\bold{M}_{ij}$.
Let $\|\bold{M}\|_{2,1} = \sum_{i=1}^{d} ||\bold{M}_{{i.}}||_{2}$ be the $\ell_{2,1}$-norm of $\bold{M}$, where
$||\bold{M}_{_{i.}}||_2=\sqrt{\sum_{j=1}^m\bold{M}_{ij}^2}$
is the $\ell_2$-norm of $\bold{M}_{_{i.}}$.
Let $||\bold{M}||_\infty= \mathop{\max}\limits_{i,j}|\bold{M}_{ij}|$
be the infinity norm of $\bold{M}$.
The trace-norm of
$\bold{M}$ is defined by $\|\bold{M}\|_{*} =
\sum_{k=1}^{rank(\bold{M})}\sigma_{k}(\bold{M})$, where
$\{\sigma_{k}(\bold{M})\}_{k=1}^{rank(\bold{M})}$ are the non-zero
singular values of $\bold{M}$ and $rank(\bold{M})$ is the rank of
$\bold{M}$. We denote $\bold{M}^T$ as the transpose of $\bold{M}$. For a vector $\bold{x}$, $||\bold{x}||_2$  represent the $\ell_2$-norm.

In the context of MTL, we assume we are given $m$ learning tasks. The $i$th ($i=1,\dots,m$) task is associated with a training set $({\mathbf{X}}^{(i)},{\mathbf{y}}^{(i)})$,
where ${\mathbf{X}}^{(i)}\in\mathbb{R}^{n_{i} \times d}$ denotes the data matrix with each row being a sample, ${\mathbf{y}}^{(i)}\in \{-1,+1\}^{n_{i}}$ denotes the target labels on $\bold{X}^{(i)}$, $d$ is the feature dimensionality, and $n_{i}$ is the number of samples for the $i$th task.
For $i=1,2,...,m$, we define $\mathbb{E}_i=\{-1,+1\}^{n_i}$ as the set of all possible $n_i$-dimension vector, each of whose elements is either $-1$ or $1$. To simplify presentation, we assume $\mathbb{E}_i=\{\bold{y}_1,\bold{y}_2,...,\bold{y}_p\}$ where $p=2^{n_i}$ and $\bold{y}_j$ is one of the possible vectors that belong to $\{-1,1\}^{n_i}$.

We define a weight matrix $\mathbf{W} = [\bold{W}_{\cdot 1}, \dots, \bold{W}_{\cdot m}]\in \mathbb{R}^{d\times m}$ on all of the $m$ tasks.
 The goal of (linear) MTL is to simultaneously learn $m$ (linear) predictors $\bold{W}_{\cdot i}\ (i=1,\dots,m)$ to minimize some loss function
 $\mathcal{L}(\bold{W}_{\cdot i};{\mathbf{X}}^{(i)},{\mathbf{y}}^{(i)})$
 (e.g. the least square loss $||{\mathbf{y}}^{(i)}-{\mathbf{X}}^{(i)} \bold{W}_{\cdot i}||_2^2$), where $\bold{W}_{\cdot i}\in\mathbb{R}^d$ is in the form of a column vector. Note that for each task, we have $\bold{X}^{(i)}=[\bold{x}_1^{(i)},\bold{x}_2^{(i)},\cdots,\bold{x}_{n_i}^{(i)}]^T$ and $\bold{y}^{(i)}=[\bold{y}_1^{(i)},\bold{y}_2^{(i)},\cdots,\bold{y}_{n_i}^{(i)}]^T$.

\section{Problem Formulations}\label{PS}
The linear MTL problem can be formulated as the generic form in (\ref{MTL}). The objective functions in many existing MTL methods are special cases of such a formulation. The following are two examples:
\begin{itemize}
\item With the regularizer ${\Omega}(\bold{W})$ being the $\ell_{2,1}$-norm $||\bold{W}||_{2,1}$ and each loss function  $\mathcal{L}(\bold{W}_{\cdot i};{\mathbf{X}}^{(i)},{\mathbf{y}}^{(i)})$  being the mean squared loss $\frac{1}{2}||\bold{y}^{(i)}-\bold{X}^{(i)}\bold{W}_{.i}||_2^2$, the problem in (\ref{MTL}) is the same as the objective used in~\cite{Liu}.
\item If the regularizer ${\Omega}(\bold{W})$ is set to be the trace-norm $||\bold{W}||_{*}$ and each loss function  $\mathcal{L}(\bold{W}_{\cdot i};{\mathbf{X}}^{(i)},{\mathbf{y}}^{(i)})$  is smooth (e.g., the mean squared loss $\frac{1}{2}||\bold{y}^{(i)}-\bold{X}^{(i)}\bold{W}_{.i}||_2^2$), the problem in (\ref{MTL}) becomes the objective used in~\cite{TraceNorm}.
\end{itemize}

The existing MTL methods mainly focus on the design of good regularizers (i.e., ${\Omega}(\bold{W})$) to catch the shared factors among multiple related tasks. The loss functions used in these methods are either to minimize the misclassified errors (for classification) or the mean squared errors (for regression). On the other hand, in practice, several evaluation metrics other than misclassified errors or mean squared errors are used the evaluation of MTL methods, such as F-score and AUC. Particularly, in the cases of MTL on imbalanced data (e.g., in a task, the number of negative samples is much larger than that of the positive samples), these metrics are more effective in performance evaluation than the standard misclassified errors or the mean squared errors.

Learning techniques of directly optimizing evaluation metrics, as known as learning with structured outputs, have been developed for many (single-task) problems, e.g., classification~\cite{Struct-SVM,SVM-Perf}, ranking~\cite{SVM-MAP}. However, despite the acknowledged importance of the metrics like F-score or AUC, little effort has been made to design MTL methods that directly optimize these evaluation metrics. The main reason is that MTL of optimizing the evaluation metrics usually results in a non-smooth objective function which is difficult to solve.

In this paper, we focus on MTL with structured outputs and propose a generic optimization procedure based on ADMM. This optimization procedure can be applied to solving a large family of MTL problems that directly optimize some evaluation metric (e.g., F-score, AUC). We call the proposed method Structured MTL (SMTL for short).

The formulation of SMTL is also a special case of (\ref{MTL}). In order to optimize some evaluation metric, we define the loss function for each task as the structured hinge loss:
\begin{displaymath}
\begin{split}
&\mathcal{L}(\bold{W}_{.i};\bold{X}^{(i)},\bold{y}^{(i)})\\
=&\mathop{\max}\limits_{{\mathbf{y}_j} \in \mathbb{E}_i} [\Delta ({{\bf{y}}^{(i)}},{\mathbf{y}_j}) - \bold{W}_{.i}^T{{\bf{X}}^{(i)}}^T({{\bf{y}}^{(i)}} - {\mathbf{y}_j})],\\
\end{split}
\end{displaymath}
where $\mathbf{y}_j$ represents any possible label assignment on $\mathbf{\mathbf{X}}^{(i)}$. $\Delta ({\mathbf{y}^{(i)}},\mathbf{y}_j)$ represents an evaluation metric to measure the distance between the true labels ${\mathbf{y}^{(i)}}$ and the other labels $\mathbf{y}_j$. For example, $\Delta(.,.)$ can be 1-F-score or 1-AUC.

The formulation of SMTL is defined as:
\begin{equation}\label{SMTL}
\begin{split}
&\min_{\bold{W}} {\Omega}(\bold{W})\\
&+\lambda \sum_{i=1}^{m}\mathop{\max}\limits_{{\mathbf{y}_j}  \in \mathbb{E}_i} [\Delta ({{\bf{y}}^{(i)}},{\mathbf{y}_j}) - \bold{W}_{.i}^T{{\bf{X}}^{(i)}}^T({{\bf{y}}^{(i)}} - {\mathbf{y}_j})].
\end{split}
\end{equation}

In this paper, we only focus on the MTL problems in the form of (\ref{SMTL}) that satisfy the following conditions:
        \begin{itemize}
            \item \textbf{Condition 1: }With respect to $\Omega \left(\mathbf{W}\right)$, there is a close-form solution for the following sub-problem
            \begin{equation}
            \begin{aligned}
            \min_{\mathbf{W}} \Omega (\mathbf{W}) + \frac{\mu }{2}\left\| {\mathbf{W}-\mathbf{M}} \right\|_F^2
            \end{aligned}
            \label{pro-op}
            \end{equation}
            where $\mathbf{M} \in {\mathbb{R}^{d  \times m }}$ and $\mu$ is a positive constant.

            \item \textbf{Condition 2: }For the evaluation metric $\Delta ({{\bf{y}}^{(i)}},{\mathbf{y}_j})$, the following sub-problem can be solve in polynomial time.
             \begin{equation}\label{cond2}
             \mathop {\argmax }\limits_{{\mathbf{y}_j}  \in \mathbb{E}_i} [\Delta ({{\bf{y}}^{(i)}},{\mathbf{y}_j}) - \bold{W}_{.i}^T{{\bf{X}}^{(i)}}^T({{\bf{y}}^{(i)}} - {\mathbf{y}_j})]
             \end{equation}
        \end{itemize}

The first condition is to restrict the regularizer $\Omega(\bold{W})$ and the second one is to restrict the evaluation metric function $\Delta(\bold{y^{(i)}},\mathbf{y}_j)$. Even under these conditions, the formulation in (\ref{SMTL}) includes a large family of MTL problems. On the one hand, for the regularizer $\Omega(\bold{W})$, the following norms that are commonly-used in MTL satisfy Condition 1:
\begin{itemize}
\item {\bf$\ell_{1,1}$-norm} For the MTL problems with $\Omega(\bold{W})=||\bold{W}||_{1,1}$, the sub-problem in (\ref{pro-op}) is known to have the close-form solution
    \begin{equation}\label{L1-close-form}
     \bold{W}=\mathcal{S}_{\frac{1}{\mu}}(\bold{M}),
    \end{equation}
    where $\mathcal{S}_{\delta}(\bold{M})=\max(\bold{M}-\delta,0)+\min(\bold{M}+\delta,0)$ is the shrinkage operator~\cite{ALM}.
\item {\bf$\ell_{2,1}$-norm} For the MTL problems with $\Omega(\bold{W})=||\bold{W}||_{2,1}$, the sub-problem in (\ref{pro-op}) is also known to have close-form solutions:
    \begin{equation}\label{L21-close-form}
\bold{W}_{j.} = \left\{ \begin{array}{ll}
\frac{||\bold{M}_{j.}||_2 - \frac{1}{\mu}}{||\bold{M}_{j.}||_2} \bold{M}_{j.} & \textrm{if $\frac{1}{\mu} <  ||\bold{M}_{j.}||_2$},\\
0 & \textrm{otherwise},
\end{array} \right.
\end{equation}
\item {\bf Trace-norm} For the MTL problems with $\Omega(\bold{W})=||\bold{W}||_{*}$, the sub-problem in (\ref{pro-op}) is also have the close-form solution by the Singular Value Threshold method~\cite{SVT}. Specifically, by assuming $\bold{U} \bold{\Sigma} \bold{V}$ be the SVD form of $\bold{M}$, the close-form solution is given by:
\begin{equation}\label{trace-close-form}
    \bold{W} = \bold{U} \mathcal{S}_{{1}/{\mu}}(\bold{\Sigma}) \bold{V}^T.
\end{equation}
\end{itemize}
On the other hand, many commonly-used metric functions satisfy the second condition. The following are two examples which we will consider in this paper:

\begin{itemize}

\item \textbf{MTL by directly optimizing F-Score} F-Score is a typical performance metric for binary classification, particularly in learning tasks on imbalanced data. F-Score is a trade-off between Precision and Recall. Specifically, given ${\mathbf{y}^{(i)}}$ and ${\mathbf{y}_j}$, we define the precision as:
    \begin{displaymath}
    Precision = \frac{\sum_{k=1}^{n_i}I({\mathbf{y}_k^{(i)}}=1\ \textit{and}\ (\mathbf{y}_j)_k=1)}{\sum_{k=1}^{n_i}I({\mathbf{y}_k^{(i)}}=1)},
    \end{displaymath}
    and the recall as:
    \begin{displaymath}
     Recall = \frac{\sum_{k=1}^{n_i}I({\mathbf{y}_k^{(i)}}=1 \ \textit{and}\ (\mathbf{y}_j)_k=1)}{\sum_{k=1}^{n_i}I({(\mathbf{y}_j)_k}=1)},
    \end{displaymath}
    where $I(\textit{condition})$ represents the indicator function that $I(\textit{condition})=1$ if $\textit{condition}$ is true, otherwise $I(\textit{condition})=0$.
    Then the F-score on ${\mathbf{y}^{(i)}}$ and ${\mathbf{y}_j}$ is defined as:
    \begin{equation}
    \begin{aligned}
    &F_\beta = \frac{(1+\beta) \times Precision \times Recall}{Precision + \beta Recall},&
    \end{aligned}
    \label{F1}
    \end{equation}
    where $\beta$ is a trade-off parameter. Hereafter, we simply set $\beta=1$. Finally, the metric function $\Delta(.,.)$ with respect to the F-score is defined by:

            \begin{equation}
                \begin{aligned}
                &\Delta ({\mathbf{y}^{(i)}},{\mathbf{y}_j}) = 1 - {F_\beta}.&
                \end{aligned}
                \label{F1_loss}
            \end{equation}
With such a form of $\Delta ({\mathbf{y}^{(i)}},{\mathbf{y}_j})$, the sub-problem in (\ref{cond2}) can be solved in polynomial time~\cite{SVM-Perf}.

\item \textbf{MTL by directly optimizing AUC} AUC is also a popular performance metric for binary classification, particularly in imbalanced learning. Given ${\mathbf{y}^{(i)}}$ and ${\mathbf{y}_j}$, the AUC metric can be calculated by :
    \begin{equation}
                \begin{aligned}
                &AUC = 1 - \frac{Swapped}{Pos\times Neg}&
                \end{aligned}
                \label{AUC}
    \end{equation}
    where $Swapped$ represents the number of ``inverted'' pairs in $\bold{y}^{(i)}$ compared to $\mathbf{y}_j$:
    \begin{displaymath}
    \begin{split}
    Swapped = &\sum_{l=1}^{n_i}\sum_{k=1}^{n_i}I(\bold{y}_l^{(i)}=1\ \textit{and}\ \bold{y}_k^{(i)}=-1)\\
    &\times I((\mathbf{y}_j)_l = -1\ \textit{and}\ (\mathbf{y}_j)_k=1).
    \end{split}
    \end{displaymath}
    $Pos$/$Neg$ represents the number of positive/negative samples in the $i$th task:
    \begin{displaymath}
    \begin{split}
    &Pos=\sum_{k=1}^{n_i}I(\bold{y}_k^{(i)}=1),\\
    &Neg=\sum_{k=1}^{n_i}I(\bold{y}_k^{(i)}=-1).\\
    \end{split}
    \end{displaymath}
    The corresponding $\Delta(.,.)$ can be defined as:

            \begin{equation}
                \begin{aligned}
                &\Delta ({\mathbf{y}^{(i)}},{\mathbf{y}_j}) = 1 - AUC.&
                \end{aligned}
                \label{AUC_loss}
            \end{equation}
With such a form of $\Delta ({\mathbf{y}^{(i)}},{\mathbf{y}_j})$, there also exists polynomial-time algorithms to solve the sub-problem in (\ref{cond2})~\cite{SVM-Perf}.

\end{itemize}
Note that here the Precision, Recall, F-Score and AUC are defined for a particular task.

\section{Proposed Optimization Procedure\label{algorithm}}
\subsection{Overview}
In this section, we present the proposed optimization procedure to solve the problem  (\ref{SMTL}). Our procedure is based on the scheme of ADMM.

For ease of presentation, we define
 \begin{displaymath}
 {\mathcal{G}_i}({\bold{W}_{.i}}) = \mathop{\max}\limits_{{\mathbf{y}_j}} [\Delta ({{\bf{y}}^{(i)}},{\mathbf{y}_j}) - \bold{W}_{.i}^T{{\bf{X}}^{(i)}}^T({{\bf{y}}^{(i)}} - {\mathbf{y}_j})],
 \end{displaymath}
  and
 \begin{displaymath}
        \mathcal{G}(\mathbf{W}) = \sum\limits_{i = 1}^m {{\mathcal{G}_i}({\bold{W}_{.i}})}.
 \end{displaymath}
Then, the problem in (\ref{SMTL}) can be re-formulated to its equivalent form in the following:
\begin{equation}\label{SMTL-ADMM}
\begin{split}
        & \min_{\bold{S},\bold{W}}\ \ \ \Omega(\mathbf{S})+\lambda \mathcal{G}(\mathbf{W}) \\
        & s.t\ \ \
        \mathbf{W} - \mathbf{S} = 0,
\end{split}
\end{equation}
where $\bold{S}\in \mathbb{R}^{d \times m}$ is an auxiliary variable.

The corresponding augmented Lagrangian function with respect to (\ref{SMTL-ADMM}) is:

%
%


        \begin{equation}\label{Lagrangian}
        \begin{split}
             \begin{array}{l}
              \mathcal{A}(\mathbf{W},\mathbf{S},\mathbf{Z})\\
             = \Omega (\mathbf{S}) + \lambda \mathcal{G}(\mathbf{W}) + \langle \mathbf{Z}, \mathbf{W} - \mathbf{S}\rangle + \frac{\mu }{2}|| {\mathbf{W} - \mathbf{S}}||_F^2\\
             \end{array}
        \end{split}
        \end{equation}
where $\mathbf{Z}$ is the Lagrangian multiplier, $\langle \cdot, \cdot\rangle$ represents the inner product of two matrices (i.e., given matrices $\bold{A}$ and $\bold{B}$, we have $\langle \bold{A}, \bold{B}\rangle=Tr(\bold{A}^T\bold{B}$), where $Tr(\bold{M})$ is the trace of the matrix $\bold{M}$), $\mu > 0$ is an adaptive penalty parameter.

Based on the ADMM scheme, the sketch of the proposed optimization procedure is shown in Algorithm 1, where in each iteration we alternatively update $\bold{W}$, $\bold{S}$ and $\bold{Z}$ by minimizing the Lagrangian function in (\ref{Lagrangian}) with other variables fixed. The update rules for $\bold{W}$, $\bold{S}$ and $\bold{Z}$ are in the following:

        \[\begin{array}{l}
        {\mathbf{S}^{\{t + 1\}}} \leftarrow \mathop {\argmin}\limits_\mathbf{S} \mathcal{A}({\mathbf{W}^{\{t\}}},\mathbf{S},{\mathbf{Z}^{\{t\} }});\\
        {\mathbf{W}^{\{t + 1\}}} \leftarrow \mathop {\argmin}\limits_\mathbf{W} \mathcal{A}(\mathbf{W},{\mathbf{S}^{\{t + 1\}}},{\mathbf{Z}^{\{t \}}});\\
        {\mathbf{Z}^{\{t + 1\}}} \leftarrow {\mathbf{Z}^{\{t\}}} + \mu ({\mathbf{W}^{\{t + 1\}}} - {\mathbf{S}^{\{t + 1\}}}).
        \end{array}\]
Note that hereafter we use $\bold{M}^{\{t\}}$ to represent the the value of variable $\bold{M}$ in the $t$-th iteration.

Next, we will present the details of solving the sub-problems with respect to $\bold{S}$ or $\bold{W}$, respectively, with other variables being fixed.

\begin{table}[htbp]
    \begin{tabular}{l}
        \hline
        {\bf Algorithm 1} The proposed ADMM procedure for \\
        the structured MTL problem (\ref{SMTL}) \\
        \hline
        \textbf{Input:}  training set $\{({\bold{X}}^{(i)},{\bold{y}}^{(i)})\}_{i=1}^{m}$, desired tolerant error $\epsilon$, \\
        \ \ \ \ \ \ \ \ \  maximal iteration number $T$. \\
        \textbf{Output:} Weight matrix $\bold{W}=[\bold{W}_{.1},\cdots,\bold{W}_{.m}]$ \\
        1. Initialize:  $\bold{Z}=\bold{S}=\bold{W} \leftarrow \bold{0}^{d\times m}$, $t \leftarrow 0$. \\
        2. Repeat: \\
        3. \ \ \   Update \\
        \ \ \ \ \ \ \ $\bold{S}^{\{t+1\}} \leftarrow \mathop {\argmin}\limits_\mathbf{S} \Omega(\mathbf{S}) + \frac{\mu }{2}|| \mathbf{S} - \mathbf{W}^{\{t\}} - \mathbf{Z}^{\{t\}} / {\mu } ||_F^2$\\
        \ \ \ \ \ \  by solving (\ref{s_l1}), (\ref{s_l21}) or (\ref{s_trace}) accordingly.\\
        4. \ \ \   For $i = 1$ to $m$ \\
        5. \ \ \ \ \ \ \ \  Update $\bold{W}^{\{t+1\}}_{.i} \leftarrow$\\ \ \ \ \ \ \ \ \ \ \ \ \ \ \ $\mathop {\argmin}\limits_{\bold{W}_{.i}} \lambda \mathcal{G}_i(\bold{W}_{.i}) + \frac{\mu}{2}|| \bold{W}_{.i} - \bold{S}_{.i}^{\{t+1\}} + \frac{\bold{Z}_{.i}^{\{t\}}}{\mu } ||_2^2$ \\
        \ \ \ \ \ \ \ \ \ \ \  by Algorithm 2. \\
        6. \ \ \   End For\\
        7. \ \ \   Update $\mathbf{Z}^{\{t+1\}} \leftarrow \mathbf{Z}^{\{t\}} + \mu(\mathbf{W}^{\{t+1\}} - \mathbf{S}^{\{t+1\}})$.\\
        8. Until $||S-W||_\infty \le \epsilon$ or $t=T$.\\
        \hline
    \end{tabular}
    \label{algorithm_admm}
\end{table}

\subsection{Solving the Sub-Problem for $\mathbf{S}$\label{pro-S}}
In the $t$-th iteration (in the outer loop) of Algorithm 1, the sub-problem for $\bold{S}$ with respect to (\ref{Lagrangian}) can be simplified as:
            \begin{equation}
            \begin{split}
            \begin{array}{l}
            \mathbf{S}^{\{t+1\}}\leftarrow \mathop{\argmin}\limits_\mathbf{S} \mathcal{A}(\mathbf{W}^{\{t\}},\mathbf{S},\mathbf{Z}^{\{t\}})\\
            = \mathop {\argmin}\limits_\mathbf{S} \Omega (\mathbf{S}) + \frac{\mu }{2}\left\| {\mathbf{W}^{\{t\}} - \mathbf{S} + \mathbf{Z}^{\{t\}} / {\mu }} \right\|_F^2\\
            \end{array}
            \label{S_update}
            \end{split}
            \end{equation}

 For different regularizer $\Omega(\bold{S})$, the solution to (\ref{S_update}) is different.
 \begin{itemize}

\item \textbf{Case 1: the $\ell_{1,1}$-norm} With $\Omega(\mathbf{S})$ being $||\mathbf{S}||_{1,1}$, by applying (\ref{L1-close-form}) to (\ref{S_update}), we have:
                \begin{equation}\label{s_l1}
                \begin{split}
                \begin{array}{l}
                {\argmin}_\mathbf{S}^{ } ||\mathbf{S}||_{1,1} + \frac{\mu }{2}\left\| {\mathbf{W}^{\{t\}} - \mathbf{S} + \mathbf{Z}^{\{t\}} / {\mu }} \right\|_F^2\\
                =
                \max(0,\mathbf{W}^{\{t\}}+ \mathbf{Z}^{\{t\}} / {\mu }-1/{\mu})\\
                +\min(0,\mathbf{W}^{\{t\}} + \mathbf{Z}^{\{t\}} / {\mu }+1/{\mu}).
                \end{array}
                \end{split}
                \end{equation}

\item \textbf{Case 2: the $\ell_{2,1}$-norm} When $\Omega(\mathbf{S})=||\mathbf{S}|{|_{2,1}}$, (\ref{S_update}) can be rewritten as:
           \begin{equation}\label{S_update_L21}
            \begin{split}
                \begin{array}{l}
                {\argmin}_\mathbf{S}^{ } ||\mathbf{S}||_{2,1} + \frac{\mu }{2}\left\| {\mathbf{W}^{\{t\}} - \mathbf{S} + \mathbf{Z}^{\{t\}} / {\mu }} \right\|_F^2.\\
                \end{array}
            \end{split}
            \end{equation}
            By applying (\ref{L21-close-form}) to (\ref{S_update_L21}), we obtain the following close-form solution:
\begin{equation}\label{s_l21}
\bold{S}_{j.} = \left\{ \begin{array}{ll}
\frac{||\bold{M}_{j.}||_2 - \frac{1}{\mu}}{||\bold{M}_{j.}||_2} \bold{M}_{j.} & \textrm{if $\frac{1}{\mu} <  ||\bold{M}_{j.}||_2$},\\
0 & \textrm{otherwise},
\end{array} \right.
\end{equation}
where $\bold{M}={\mathbf{W}^{\{t\}}+ \mathbf{Z}^{\{t\}} / {\mu }}$.
  \item\textbf{Case 3: the trace-norm} When $\Omega(\mathbf{S})=||\mathbf{S}|{|_{*}}$, we can apply (\ref{trace-close-form}) to (\ref{S_update}) and obtain the following close-form solution:
            \begin{equation}
            \begin{split}
            \begin{array}{l}
                {\argmin}_\mathbf{S}^{ } ||\mathbf{S}||_{*} + \frac{\mu }{2}\left\| {\mathbf{W}^{\{t\}} - \mathbf{S} + \mathbf{Z}^{\{t\}} / {\mu }} \right\|_F^2.\\
                =\bold{U}(\max(0,\mathbf{\Sigma}-1/\mu)+\min(0,\mathbf{\Sigma}+1/\mu)) \bold{V}^T,
                \end{array}
            \end{split}
            \label{s_trace}
            \end{equation}
            where $\mathbf{U}\mathbf{\Sigma} \mathbf{V}^T$ is the SVD form of ${\mathbf{W}^{\{t\}} + \mathbf{Z}^{\{t\}} / {\mu }}$.

\end{itemize}

\subsection{Solving the Sub-Problem for $\mathbf{W}$\label{pro-W}}
\subsubsection{Formulation}
In the $t$-th outer iteration in Algorithm 1, the sub-problem for $\bold{W}$ with respect to (\ref{Lagrangian}) can be reformulated as:

\begin{equation}
            \begin{split}
            \begin{array}{l}
            \mathbf{W}^{\{t+1\}}\leftarrow \mathop{\argmin}\limits_\mathbf{W} \mathcal{A}(\mathbf{W},\mathbf{S}^{\{t+1\}},\mathbf{Z}^{\{t\}})\\
            = \mathop {\argmin}\limits_\mathbf{W} \lambda\mathcal{G}(\mathbf{W}) + \frac{\mu }{2}\left\| {\mathbf{W} - \mathbf{S}^{\{t+1\}} + \mathbf{Z}^{\{t\}} / {\mu }} \right\|_F^2\\
            = \mathop {\argmin}\limits_\mathbf{W} \sum_{i=1}^{m}\lambda\mathcal{G}_i(\mathbf{W}_{.i}) + \frac{\mu }{2}\left\| {\mathbf{W}_{.i} - \mathbf{S}_{.i}^{\{t+1\}} + \mathbf{Z}_{.i}^{\{t\}} / {\mu }} \right\|_F^2\\
            \end{array}
            \end{split}
            \label{W_sub_problem}
\end{equation}

 To simplify presentation, we denote $\bold{b}_i=\mathbf{S}_{.i}^{\{t+1\}} - \mathbf{Z}_{.i}^{\{t\}} / {\mu }$. Then, the problem in (\ref{W_sub_problem}) can be separated into $m$ independent sub-tasks:

\begin{equation}
            \begin{split}
            \begin{array}{l}
            \mathop {\min}\limits_{\mathbf{W}_{.i}} \lambda\mathcal{G}_i(\mathbf{W}_{.i}) + \frac{\mu }{2}\left\| {\mathbf{W}_{.i} - \mathbf{b}_{i}} \right\|_F^2, i=1,...,m.\\
            \end{array}
            \end{split}
            \label{w_sub_problem}
\end{equation}

For $j=1,2,...,p$, we define $\mathbf{K}=[\bold{K}_{.1},\bold{K}_{.2},...,\bold{K}_{.p}]$ with $\bold{K}_{.j}={\mathbf{X}^{(i)}}^T(\mathbf{y}^{(i)}-\bold{y}_j)+\frac{\mu}{\lambda}\bold{b}_i$, $\mathbf{\Delta}=(\mathbf{\Delta}_1,\mathbf{\Delta}_2,...,\mathbf{\Delta}_p)^T$ with $\mathbf{\Delta}_j=\Delta ({\mathbf{y}^{(i)}},{{\mathbf{y}}_j})$. Then, the problem in (\ref{w_sub_problem}) can be simplified as:
\begin{equation}
            \begin{split}
            \begin{array}{l}
            \mathop {\min}\limits_{\mathbf{W}_{.i}} \lambda\mathcal{G}_i(\mathbf{W}_{.i}) + \frac{\mu }{2}\left\| {\mathbf{W}_{.i} - \mathbf{b}_{i}} \right\|_F^2\\
            = \mathop {\min}\limits_{\mathbf{W}_{.i}} \frac{\mu }{2}(|| \mathbf{W}_{.i}||_F^2+||\mathbf{b}_{i}||_F^2-2\mathbf{W}_{.i}^T\mathbf{b}_{i})\\
            +\lambda \mathop{\max}\limits_{\bold{y}_j\in \mathbb{E}_i} [\Delta ({{\bf{y}}^{(i)}},{\bold{y}_j}) - \mathbf{W}_{.i}^T{{\bf{X}}^{(i)}}^T({{\bf{y}}^{(i)}} - {\bold{y}_j})].\\
            \end{array}
            \end{split}
            \label{wi_struct_org}
\end{equation}
By re-scaling the objective (\ref{wi_struct_org}) by $\mu$ and drop the terms independent of $\mathbf{W}_{.i}$ and $\bold{y}_j$, we have:
\begin{equation}
            \begin{split}
            \begin{array}{l}
            \mathop {\min }\limits_{\mathbf{W}_{.i}} \frac{1}{2}||{\mathbf{W}_{.i}}||_2^2 + \frac{\lambda}{\mu }\mathop {\max }\limits_j [{\mathbf{\Delta}_j} - {({\mathbf{W}_{.i}^T}\mathbf{K})_{j}}]\\
            \end{array}
            \end{split}
            \label{wi_struct}
\end{equation}

The existence of the max operator on exponential number of elements makes it difficult to optimize the objective in (\ref{wi_struct}). To tackle this issue, in the next two subsection, we derive the Fenchel dual~\cite{Fenchel} form of (\ref{wi_struct}) and develop a coordinate ascent algorithm to solve this dual form.
\subsubsection{Fenchel Dual Form of (\ref{wi_struct})}
In this subsection, we derive the Fenchel dual~\cite{Fenchel} form of (\ref{wi_struct}). To simplify presentation, we use $\mathbf{w}$ to represent $\mathbf{W}_{.i}$. Then we re-formulate the primal form in (\ref{wi_struct}) as:
\begin{equation}\label{primal}
\begin{split}
\mathop {\min }\limits_{\mathbf{w}} \mathcal{P}(\bold{w})=\mathcal{M}(\bold{w})+\mathcal{N}(-\mathbf{w}^T\mathbf{K})\\
=\frac{1}{2}||{\mathbf{w}}||_2^2 + \frac{\lambda}{\mu }\max_{j}^{ }(\mathbf{\Delta}^T - \mathbf{w}^T\mathbf{K})_j,\\
\end{split}
\end{equation}
where we define $\mathcal{M}(\bold{w})=\frac{1}{2}||{\mathbf{w}}||_2^2$ and $\mathcal{N}(-\mathbf{w}^T\mathbf{K})=\frac{\lambda}{\mu }\max_{j}^{ }(\mathbf{\Delta}^T - \mathbf{w}^T\mathbf{K})_j$.

Before deriving the dual form of (\ref{primal}), we first introduce the definition (Definition \ref{def1}) and the main properties (Theorem \ref{fenchel-young} and \ref{fenchel dual}) of Fenchel duality.

\begin{definition} \label{def1}
	The Fenchel conjugate of function $f(\boldsymbol{x})$ is defined as $f^{*}(\boldsymbol{\theta})= \max_{\boldsymbol{x} \in dom (f)} (\langle \boldsymbol{\theta},\boldsymbol{x} \rangle
	-f(\boldsymbol{x}))$.
\end{definition}
\begin{theorem}
	(Fenchel-Young inequality:~\cite{c7},
	Proposition 3.3.4) Any points $\boldsymbol{\theta}$ in the domain of function $f^{*}$ and $\boldsymbol{x}$ in the domain of
	function $f$ satisfy the inequality:
	\begin{equation}
	f(\boldsymbol{x}) + f^{*}(\boldsymbol{\theta}) \geq \langle \boldsymbol{\theta},\boldsymbol{x} \rangle
	\label{fenchel_eqt}
	\end{equation}
	The equality holds if and only if $\boldsymbol{\theta} \in \partial f(\boldsymbol{x})$.
	\label{fenchel-young}
\end{theorem}

\begin{theorem}
	(Fenchel Duality inequality, see e.g.,Theorem 3.3.5 in~\cite{c7}) Let $\mathcal{M}:\mathbb{R}^{d} \rightarrow (-\infty,+\infty]$ and $\mathcal{N}:\mathbb{R}^{p} \rightarrow (-\infty,+\infty]$ be two closed and convex functions, and $\mathbf{K}$ be a $\mathbb{R}^{d \times p}$ matrix. Then we have
	\begin{equation}\label{primal-dual}
	\sup_{\boldsymbol{\alpha}} -\mathcal{M}^{*}(\mathbf{K}\boldsymbol{\alpha}) - \mathcal{N}^{*}(\boldsymbol{\alpha}) \leq \inf_{\boldsymbol{w}} \mathcal{M}(\boldsymbol{w}) + \mathcal{N}(-\boldsymbol{w}^T\mathbf{K}),
	\end{equation}
	where $\boldsymbol{\alpha}\in \mathbb{R}^{p}$ and $\boldsymbol{w}\in \mathbb{R}^{d}$. The equality holds if and only if $0 \in (dom(\mathcal{N}) - \mathbf{K}^T dom(\mathcal{M}))$.\
	\label{fenchel dual}
\end{theorem}
Note that the right hand side of the inequality in (\ref{primal-dual}) is called the primal form and the left hand side of (\ref{primal-dual}) is the corresponding dual form.

With Definition \ref{def1}, it is known (see, e.g., ~\cite{Boosting}, Appendix B) that the Fenchel dual norm (i.e., the Fenchel conjugate) of the $\ell_2$-norm $f(\bold{x})=\frac{1}{2}||\bold{x}||_2^2$ is also the $\ell_2$-norm $f^*(\bold{\theta})=\frac{1}{2}||\bold{\theta}||_2^2$. Hence, the Fenchel conjugate of $\mathcal{M}(\bold{w})=\frac{1}{2}||\bold{w}||_2^2$ is
\begin{equation}
\begin{split}
\mathcal{M}^*(-\bold{K\alpha})=\frac{1}{2}||\bold{-K\alpha}||_2^2
\end{split}
\label{M*}
\end{equation}

It is known (~\cite{Boosting}, Appendix B) that the Fenchel conjugate of $f(\bold{x}+\bold{y})$ is $f^*(\bold{\theta})-\langle \bold{\theta},\bold{y}\rangle$, the Fenchel conjugate of $cf(\bold{x})$ ($c>0$) is $cf^{*}(\bold{\theta}/c)$. Then we can derive that the Fenchel conjugate of $cf(\bold{x}+\bold{y})$ is
\begin{equation}\label{fenchel_dual_complex}
cf^{*}(\bold{\theta}/c)-\langle \bold{\theta},\bold{y}\rangle.
\end{equation}

In addition, the Fenchel conjugate of $f(\bold{x})=\max_j^{ }(\bold{x}_j)$ is $I_{\theta_i\geq 0, \sum_i \theta_i=1}(\theta)$ with $I_{condition}(.)$ being the indicator function that $I_{condition}(\theta)=0$ if $condition$ is true and otherwise $I_{condition}(\theta)=+\infty$ (see~\cite{Boosting}, Appendix B). For convenience, we denote $\mathcal{Q}(\bold{x})=\max_j^{ } (\bold{x}_j)$. It is easy to verify that $\mathcal{N}(-\mathbf{w}^T\mathbf{K})=\frac{\lambda}{\mu }\max_{j}^{ }(\Delta^T - \mathbf{w}^T\mathbf{K})_j=\frac{\lambda}{\mu}\mathcal{Q}(\Delta^T-\mathbf{w}^T\mathbf{K})$. Hence, by using (\ref{fenchel_dual_complex}), the Fenchel conjugate of $\mathcal{N}(-\mathbf{w}^T\mathbf{K})$ is:

\begin{equation}
\begin{split}
&\mathcal{N}^*(\mathbf{\alpha})=\frac{\lambda}{\mu}\mathcal{Q}^*((\bold{\alpha})/(\frac{\lambda}{\mu}))-\langle \mathbf{\alpha},\bold{\Delta}\rangle\\
&= \left\{ \begin{array}{l}
- {\mathbf{\Delta}^T}\mathbf{\alpha} ,\ \sum\limits_{k = 1}^p {\mathbf{\alpha}_k  = \frac{\lambda}{\mu }\ and\  \mathbf{\alpha}_k  \ge 0,\ k = 1, \cdots ,p} ;\\
+ \infty ,\ \ \ \ otherwise.
\end{array} \right.
\end{split}
\label{N*}
\end{equation}

With (\ref{M*}), (\ref{N*}) and (\ref{primal-dual}), we have that the dual form of (\ref{primal}) is:
\begin{equation}
\begin{split}
&\mathop {\max }\limits_{\mathbf{\alpha}}\mathcal{D}(\mathbf{\alpha}) \\
&=\mathop {\max }\limits_{\mathbf{\alpha}} -\mathcal{M}^*(\mathbf{K}\alpha) - \mathcal{N}^*(\alpha)\\
&=\mathop {\max }\limits_{\mathbf{\alpha}}   - \frac{1}{2}{\mathbf{\alpha}^T}{\mathbf{K}^T}\mathbf{K}\mathbf{\alpha}  + \mathbf{\Delta}^T\mathbf{\alpha} \\
&s.t.\ \sum\limits_{k = 1}^p \mathbf{\alpha}_k  = \frac{\lambda}{\mu }\ and\  \mathbf{\alpha}_k  \ge 0,\ k = 1, \cdots ,p\\
\end{split}
\label{dual}
\end{equation}

The dual form in (\ref{dual}) is a smooth quadratic function with linear constraints, which is easier to be optimized compared to its primal form in (\ref{primal}).

\subsubsection{Primal-Dual Algorithm via Coordinate Ascent}
In this subsection, we develop a coordinate ascent algorithm to optimize the objective in (\ref{dual}), where we use the primal-dual gap $\mathcal{P}(\bold{w})-\mathcal{D}(\bold{\alpha})$ as the early stopping criterion. Coordinate ascent is a widely-used method in various machine learning problems (e.g.,~\cite{liblinear,Boosting,Lai-TC,CNNH}).

\begin{table}[htbp]
	\begin{tabular}{l}
		\hline
		{\bf Algorithm 2} Primal-dual algorithm via coordinate ascent\\
		\hline
		\textbf{Input}: $\mathbf{b}_i$, $\epsilon_F$, $\lambda$, $\mu$,
		maximal iteration number $T_F$ \\
		\textbf{Output}: $\bold{w}$\\
		1.\ Initialize: $v\leftarrow 0$, $\bold{\hat{w}}\leftarrow 0$ \\
		2.\ Repeat: \\
		3.\ \ \ \ Find the largest element $(g_\alpha)_j$ in the gradient vector \\
		\ \ \ \ \ \ $g_\alpha=\nabla \mathcal{D}(\alpha)$ by solving (\ref{find_most_violated_contraint}) via Algorithm 3 for F-score \\
		\ \ \ \ \ \ (or Algorithm 4 for AUC). \\
		4.\ \ \ \ $\mathbf{\Delta}_j \leftarrow \Delta ({\mathbf{y}^{(i)}},{{\mathbf{y}}_j})$\\
		5.\ \ \ \ $\bold{K}_{.j} \leftarrow {\mathbf{X}^{(i)}}^T(\mathbf{y}^{(i)}-\bold{y}_j)+\frac{\mu}{\lambda}\bold{b}_i$\\
		6.\ \ \ \ Calculate $\gamma$ by (\ref{0-1-simplified}).\\
		7.\ \ \ \ Update $\bold{\hat{w}}$ by (\ref{w}).\\
		8.\ \ \ \ Update $v$ by (\ref{alphadelta})\\
		9.\ Until $\bold{\hat{w}}^T\bold{\hat{w}}+\max_j^{ }(g_\alpha)_j\leq \epsilon_F$ or iteration number reaches $T_F$ \\
		10.\ $\bold{w}\leftarrow \bold{\hat{w}}$\\
		\hline
	\end{tabular}
	\label{coordinate_ascent}
\end{table}

The proposed coordinate ascent algorithm is shown in Algorithm 2. Next, we sketch the main steps the proposed algorithm in the following:

\
\\
\textbf{Repeat}
\begin{itemize}
	\item Select an index $j$ with the $j$-th element $(\nabla_\alpha\mathcal{D}(\alpha))_j$ in the gradient vector $\nabla_\alpha\mathcal{D}(\alpha)$ having the largest element.
	\item Update $\alpha_j$ with other $\alpha_k$ ($k\neq j$) fixed, in a manner of greedily increasing the value of $\mathcal{D}(\alpha)$.
\end{itemize}
\textbf{Until} the early stopping criterion $\mathcal{P}(\bold{w})-\mathcal{D}(\bold{\alpha})\le \epsilon_F$ is satisfied.

\

In each iteration, the proposed algorithm has three main building blocks:

\
\\
\textbf{The First Step} is to select an index $j$ that the $j$-th element is the largest element in the gradient vector for the dual objective $\mathcal{D}({\alpha})$. Specifically, the gradient vector with respect to $\alpha$ for $\mathcal{D}({\alpha})$ is:
\begin{displaymath}
g_\alpha=\nabla_{\alpha} \mathcal{D}({\alpha})=-\mathbf{K}^T\mathbf{K}\alpha+\mathbf{\Delta},
\end{displaymath}
and the largest element in $\nabla_{\alpha} \mathcal{D}({\alpha})$ is:
\begin{displaymath}
\begin{split}
(g_\alpha)_j=(\nabla_{\alpha} \mathcal{D}({\alpha}))_j=\max_j^{ }\mathbf{\Delta}_j-(\mathbf{K}\alpha)^T\mathbf{K}_{.j}.
\end{split}
\end{displaymath}
We denote $\bold{\hat{w}}=\mathbf{K}\alpha$. Then, with the definition of $\mathbf{\Delta}_j$ and $\bold{K}_{.j}$, we have:
\begin{equation}\label{find_most_violated_contraint}
\begin{split}
(\nabla_{\alpha} \mathcal{D}({\alpha}))_j
=\max_j^{ }\Delta({\mathbf{y}^{(i)}},{\bold{y}_j})-\mathbf{\hat{w}}^T{\mathbf{X}^{(i)}}^T({\mathbf{y}^{(i)}} - {\bold{y}_j}).\\
\end{split}
\end{equation}

Interestingly, the problem in (\ref{find_most_violated_contraint}) is essentially the same as the problems of ``finding the most violated constraint'' in Structured-SVMs (e.g., the problem (7) in~\cite{SVM-Perf}). For several commonly-used evaluation metrics $\Delta(.,.)$, efficient algorithm in polynomial-time were proposed to solve the problems of ``finding the most violated constraint''. One can directly use these inference algorithms to solve (\ref{find_most_violated_contraint}) of selecting the largest element from the gradient vector $\nabla_{\alpha} \mathcal{D}({\alpha})$. For example, when $\Delta(.,.)$ corresponds to F-score, one can use Algorithm 2 in~\cite{SVM-Perf} to solve (\ref{find_most_violated_contraint}); when $\Delta(.,.)$ corresponds to AUC, one can use Algorithm 3 in~\cite{SVM-Perf} to solve (\ref{find_most_violated_contraint}). For self-containness, we shown these two algorithms with our notations in Algorithm 3 and 4. Note that Algorithm 3 and 4 have the time complexity in $O(n^2_i)$ and $O(n_i \log n_i)$, respectively.

%
%
%
%

\begin{table}[htbp]
	\begin{tabular}{l}
		\hline
		{\bf Algorithm 3} Algorithm to solve (\ref{find_most_violated_contraint}) with
		loss function \\ defined on F-score\\
		\hline
		\textbf{Input}: $n=n_i, {\mathbf{X}^{(i)}}=(\mathbf{x}^{(i)}_1,\ldots,\mathbf{x}^{(i)}_{n})^T$,\\
		\ \ \ \ \ \ \ \ \  $\mathbf{y}^{(i)}=(\mathbf{y}^{(i)}_1,\ldots,\mathbf{y}^{(i)}_{n})^T$, $\mathbf{w}$\\
		\textbf{Output}: $\mathbf{y}_j$\\
		1.\ Initialize: $(k^p_1,\ldots,k^p_{Pos})\leftarrow sort\{k : \mathbf{y}^{(i)}_k = 1\}$ by $ \mathbf{w}^T\mathbf{x}^{(i)}_k$\\
		\ \ \ \ \ \ \ \ \ \ \ \ \ \ \ \  $(k^n_1,\ldots,k^n_{Neg})\leftarrow sort\{k : \mathbf{y}^{(i)}_k = -1\}$ by $ \mathbf{w}^T\mathbf{x}^{(i)}_k$\\
		2.\ For  $a \in [0, \ldots, Pos] $ do: \\
		3.\ \ \ \ $c\leftarrow Pos - a$\\
		4.\ \ \ \ Set $l{_{k_1^p}}, \ldots ,l{_{k_a^p}}$ to $1$ and set $l{_{k_{a + 1}^p}}, \ldots ,l{_{k_{Pos}^p}}$ to $-1$\\
		5.\ \ \ \ For  $d\in [0, \ldots,Neg] $ do: \\
		6.\ \ \ \ \ \ \ \ $b\leftarrow Neg - d$\\
		7.\ \ \ \ \ \ \ \ Set $l{_{k_1^n}}, \ldots ,l{_{k_b^n}}$ to $1$ and set $l{_{k_{b + 1}^n}}, \ldots ,l{_{k_{Neg}^n}}$ to $-1$\\
		8.\ \ \ \ \ \ \ \ $v \leftarrow  \Delta ({{\bf{y}}^{(i)}}, (l_1,\ldots,l_n)^T) + {\mathbf{w}^T}\sum\limits_{k = 1}^n {l{_k}{\mathbf{x}^{(i)}_k}} $,\\
		\ \ \ \ \ \ \ \ \ \ where $\Delta(\cdot,\cdot)$ is defined by (\ref{AUC_loss})\\
		9.\ \ \ \ \ \ \ \ If $v$ is the largest so far, then:\\
		10.\ \ \ \ \ \ \ \ \ \ $\mathbf{y}_j \leftarrow (l_1,\ldots,l_n)^T$\\
		11.\ \ \ \ \ \ \ \ End if\\
		12.\ \ \ \ End for\\
		13.\ End for\\
		\hline
	\end{tabular}
	\label{inf_f1}
\end{table}

\begin{table}[htbp]
	\begin{tabular}{l}
		\hline
		{\bf Algorithm 4} Algorithm to solve (\ref{find_most_violated_contraint}) with
		loss function \\ defined on AUC\\
		\hline
		\textbf{Input}: $n=n_i, {\mathbf{X}^{(i)}}=(\mathbf{x}^{(i)}_1,\ldots,\mathbf{x}^{(i)}_{n})^T$,\\
		\ \ \ \ \ \ \ \ \  $\mathbf{y}^{(i)}=(\mathbf{y}^{(i)}_1,\ldots,\mathbf{y}^{(i)}_{n})^T$, $\mathbf{w}$\\
		\textbf{Output}: $\mathbf{y}_j$\\
		1.\ Initialize: for $k \in \{ k:{\bf{y}}_k^{(i)} = 1\}$ do ${q_k} \leftarrow   - 0.25 + {\mathbf{w}^T}{\mathbf{x}^{(i)}_k}$\\
		\ \ \ \ \ \ \ \ \ \ \ \ \ \ \ \  for $k \in \{ k:{\bf{y}}_k^{(i)} = -1\}$ do ${q_k} \leftarrow    0.25 + {\mathbf{w}^T}{\mathbf{x}^{(i)}_k}$\\
		2.\ $({r_1}, \ldots ,{r_n}) \leftarrow  $ sort $\{ 1, \ldots ,n\}$ by ${q_k}$ \\
		3.\ ${q_{Pos}} \leftarrow Pos$, $q_{Neg} \leftarrow 0$\\
		4.\  For  $k \in [1, \ldots, n] $ do: \\
		5.\ \ \ \ If $\mathbf{y}^{(i)}_{r_k} >0$, then:\\
		6.\ \ \ \ \ \ \ ${l_{{r_k}}} \leftarrow  (Neg - 2{q_n})$\\
		7.\ \ \ \ \ \ \ ${q_{Pos}} \leftarrow  {q_{Pos}} - 1$\\
		8.\ \ \ \ else \\ 
		9.\ \ \ \ \ \ \ ${l_{{r_k}}} \leftarrow  ( - Pos + 2{q_{Pos}})$\\
		10.\ \ \ \ \ \ ${q_{Neg}} \leftarrow  {q_{Neg }} + 1$\\
		11.\ \ \ \ End if\\
		12.\ End for\\
		13.\ Convert $(l_1,\ldots,l_n)$ to $\mathbf{y}_j$ according to some \\
		\ \ \ \ threshold value.\\
		\hline
	\end{tabular}
	\label{inf_auc}
\end{table}
\
\\
\textbf{The Second Step} is to update $\alpha_j$ by fixing other variable $\alpha_k (k\ne j)$, given the selected index $j$.

We define the update rules for $\alpha$ as:
\begin{equation}
\alpha \leftarrow (1-\gamma)\alpha + \frac{\gamma\lambda}{\mu} e_j,
\label{alpha}
\end{equation}
where $0\le \gamma \le 1$ and $e_j$ denotes the $n_i$-dimension vector with the $j$-th element being one and other elements being zeros. It is worth noting that, given $\alpha_j\ge 0$ and $\sum_j\alpha_j=\lambda/\mu$ before updating, and $0\le\gamma\le 1$, this form of rules in (\ref{alpha}) guarantees that $\alpha_j\ge 0$ and $\sum_j\alpha_j=\lambda/\mu$ still hold after updating.

By substituting (\ref{alpha}) into (\ref{dual}), we obtain the corresponding optimization problem with respect to $\gamma$:

\begin{equation}
\begin{split}
\begin{array}{l}
\mathop {\max }\limits_\gamma   - \frac{1}{2}{[(1 - \gamma )\alpha  + \frac{\gamma\lambda }{\mu }{e_j}]^T}{\mathbf{K}^T}\mathbf{K}[(1 - \gamma )\alpha  + \frac{\gamma\lambda }{\mu }{e_j}]\\
\ \ \ \ \ \ + {[(1 - \gamma )\alpha  + \frac{\gamma\lambda }{\mu }{e_j}]^T}\mathbf{\Delta}
\end{array}
\end{split}
\label{gama-obj}
\end{equation}

Intuitively, our goal is to find $\gamma\in [0,1]$ to increase the dual objective $\mathcal{D}(\alpha)$ as much as possible. By setting the gradient of (\ref{gama-obj}) with respect to $\gamma$ to zero, we have
\begin{displaymath}
\begin{array}{l}
||\mathbf{K}({e_j}\lambda/\mu  - \alpha )||_2^2\gamma  + {({e_j}\lambda/\mu  - \alpha )^T}{\mathbf{K}^T}\mathbf{K}\alpha \\
- {({e_j}\lambda/\mu  - \alpha )^T}\mathbf{\Delta}  = 0
\end{array}
\end{displaymath}
By simple algebra, we have
\begin{equation}
\begin{aligned}
\begin{array}{l}
\gamma  =  - \frac{{{{({e_j}\lambda/\mu  - \alpha )}^T}({\mathbf{K}^T}\mathbf{K}\alpha  - \mathbf{\Delta} )}}{{||\mathbf{K}({e_j}\lambda/\mu  - \alpha )||_2^2}}\\
\end{array}
\end{aligned}
\label{gama}
\end{equation}
To ensure that $0\le\gamma\le 1$, we make further restriction on $\gamma$:

\begin{equation}
\begin{aligned}
\gamma =\max(\min(- \frac{{{{({e_j}\lambda/\mu  - \alpha )}^T}({\mathbf{K}^T}\mathbf{K}\alpha  - \mathbf{\Delta} )}}{{||\mathbf{K}({e_j}\lambda/\mu  - \alpha )||_2^2}},1),0)
\end{aligned}
\label{0-1}
\end{equation}

The calculation of $\gamma$ in (\ref{0-1}) depends on the calculation of $\bold{K\alpha}$ and $\alpha^T\mathbf{\Delta}$. However, since $\bold{K}\in \mathbb{R}^{d\times p}$, $\mathbf{\Delta},\alpha \in \mathbb{R}^{p}$ and $p=2^{n_i}$, the time of directly calculating either $\bold{K\alpha}$ or $\alpha^T\mathbf{\Delta}$ depends exponentially on $n_i$, which may often unaffordable. In order to improve efficiency, we maintain auxiliary variable to reduce the computation cost. Remind that we have defined $\bold{\hat{w}}=\bold{K\alpha}$. We also define ${v}=\alpha^T\mathbf{\Delta}$. We maintain $\bold{\hat{w}}$ and $v$ during the iterations.

With the update rule (\ref{alpha}) for $\bold{\alpha}$, we can easily derive the corresponding update rules for $\bold{\hat{w}}$ and $\bold{v}$, respectively:
\begin{equation}
\begin{aligned}
\bold{\hat{w}} \leftarrow (1 - \gamma)\bold{\hat{w}} +\frac{\gamma\lambda}{\mu}\bold{K}_{.j},
\end{aligned}
\label{w}
\end{equation}

%
\begin{equation}
\begin{aligned}
{v} \leftarrow (1-\gamma){v}+\frac{\gamma\lambda}{\mu}\mathbf{\Delta}_j.
\end{aligned}
\label{alphadelta}
\end{equation}
Obviously, the update rule for $\bold{\hat{w}}$ (or $v$) has the time complexity $O(d)$ (or $O(1)$).

With the maintained $\bold{\hat{w}}$ and $v$, the update rule in (\ref{0-1}) can be simplified to:
\begin{equation}
\begin{aligned}
\gamma \leftarrow \max(\min(- \frac{{\frac{\lambda}{\mu}(\bold{K}_{j.}^T\bold{\hat{w}}-\mathbf{\Delta}_j)-\bold{\hat{w}}^T\bold{\hat{w}}+v  }}{{||\frac{\lambda}{\mu}\bold{K}_{.j}  - \bold{\hat{w}} )||_2^2}},1),0),
\end{aligned}
\label{0-1-simplified}
\end{equation}
where the time complexity of update $\gamma$ in (\ref{0-1-simplified}) is reduced to $O(d)$.

\
\\
\textbf{The early stopping criterion} is defined based on the primal-dual gap $\mathcal{P}(\bold{w})-\mathcal{D}(\bold{\alpha})\leq \epsilon_F$ where the parameter $\epsilon_F$ is the pre-defined tolerance. Assume $\mathcal{P}(\bold{w}^{\star})$ is the optimal value of the primal objective (\ref{primal}). According to Theorem \ref{fenchel dual}, we have:
\begin{displaymath}
\mathcal{P}(\bold{w})-\mathcal{P}(\bold{w}^{\star})\leq\mathcal{P}(\bold{w})-\mathcal{D}(\bold{\alpha})\leq \epsilon_F.
\end{displaymath}

It is worth noting that, by using the update rule ({\ref{alpha}) with $0\leq \gamma \leq 1$, Algorithm 2 guarantees that $\alpha$ satisfies the constraints $\alpha_k\ge 0$ and $\sum_k \alpha_k=\lambda/\mu$ in all of the iterations. In order words, we have $\mathcal{N}^*(\alpha)<\infty$ in all of the iterations. Hence, with (\ref{primal}) and (\ref{dual}), we have:
	\begin{equation}\label{gap-org}
	\begin{split}
	&\mathcal{P}(\bold{w})-\mathcal{D}(\bold{\alpha})\\
	 &=\mathcal{M}(\bold{w})+\mathcal{M}^*(\bold{K\alpha})+\mathcal{N}(-\mathbf{w}^T\mathbf{K})+\mathcal{N}^*(\alpha)
	\end{split}
	\end{equation}
	
	With Theorem \ref{fenchel-young}, we have $\mathcal{M}(\bold{w})+\mathcal{M}^*(\bold{K\alpha})\ge \langle \bold{w},\bold{K\alpha}\rangle$, where the equality holds when $\bold{w}=\bold{K\alpha}=\bold{\hat{w}}$. In order to greedily upper-bounded the gap $\mathcal{D}(\bold{\alpha}^{\star})-\mathcal{D}(\bold{\alpha})$, we set $\bold{w}=\bold{K\alpha}=\bold{\hat{w}}$ in (\ref{gap-org}) and obtain:
	
	\begin{equation}\label{gap}
	\begin{split}
	&\mathcal{P}(\bold{w})-\mathcal{D}(\bold{\alpha})\\
	&=\langle \bold{\hat{w}},\bold{K\alpha}\rangle+\mathcal{N}(\mathbf{\hat{w}}^T\mathbf{K})+\mathcal{N}^*(\alpha)\\
	&=\bold{\hat{w}}^T\bold{\hat{w}}+\max_j^{ }(g_\alpha)_j-v
	\end{split}
	\end{equation}
	
	Consequently, the early stopping criterion is set to be
	$\bold{\hat{w}}^T\bold{\hat{w}}+\max_j^{ }(g_\alpha)_j-v\leq \epsilon_F$, which can be calculated in time $O(d)$.

\subsection{Convergence Analysis}
 For the sub-problem w. r. t. $\mathbf{W}$ (see Section \ref{pro-W}), the proposed coordinate ascent method is similar to those in~\cite{Boosting,Lai-TC}. By using similar proof techniques to those of~\cite{Boosting,Lai-TC} (e.g., see the proofs of Theorem 1 in~\cite{Lai-TC}), we can derive that, after $T$ iteration in Algorithm 2, we have $\mathcal{D}(\bold{\alpha}^{\star})-\mathcal{D}(\bold{\alpha})\leq\mathcal{P}(\bold{w})-\mathcal{D}(\bold{\alpha})\leq \epsilon_F=O(\frac{1}{T})$. Note that $\mathcal{D}(\bold{\alpha}^{\star})=\mathcal{P}(\bold{w}^{\star})$, where $\mathcal{D}(\bold{\alpha}^{\star})$ and $\mathcal{P}(\bold{w}^{\star})$ are the optimal solution of (\ref{dual}) and ({\ref{primal}}) respectively. Ideally, for all the tasks, if we set the iteration number $T$ to be sufficient large, we can solve the sub-problem w,r.t. $W$ exactly (by ignoring the small numerical errors).

In addition, as discussed in Section \ref{pro-S}, the sub-problems w. r. t. $\mathbf{S}$ can be solved exactly by closed-form solutions. Hence, the objective (\ref{SMTL-ADMM}) is convex subject to linear constraints, and both of its subproblems can be solved exactly. Based on existing theoretical results~\cite{ADMM,He2012}, we have that Algorithm 1 converges to global optima with a $O(1/\epsilon)$ convergence rate.

\section{Experiments}
\subsection{Overview}
In this section, we evaluate and compare the performance of the proposed SMTL method on several benchmark datasets. For the regularizer $\Omega(\bold{S})$ in (\ref{SMTL-ADMM}), we consider $||\bold{S}||_{1,1}$, $||\bold{S}||_{2,1}$ and $||\bold{S}||_{*}$, respectively. For the evaluation metric $\Delta(.,.)$ used in $\mathcal{G}(\bold{W})$ in (\ref{SMTL-ADMM}), we consider $F_1$-score (with $\beta=1$) and AUC. These settings lead to six variants of SMTL.

Here we focus on MTL for classification. Given a specific regularizer (i.e., $||\bold{S}||_{1,1}$, $||\bold{S}||_{2,1}$ or $||\bold{S}||_{*}$), we choose these methods as baselines: (1) single-task structured SVM that directly optimizes AUC (StructSVM)~\cite{SVM-Perf}, we train it on each of the individual tasks and average the results. (2) MTL with hinge loss for classification (MTL-CLS). (3) MTL with least squares loss for regression (MTL-REG). (4) RAkEL, a meta algorithm using random $k$-label sets~\cite{Tsoumakas}. (5) MLCSSP, a method spanning the original label space by subset of labels~\cite{James}. (6) AdaBoostMH, a method based on AdaBoost~\cite{schapire2000boostexter}. (7) HOMER, a method based on the hierarchy of multi-label learners~\cite{Vlahavas}. (8) BR, the binary relevance method~\cite{LP-BR}. (9) LP, the label power-set method~\cite{LP-BR}. (10) ECC, the ensembles of classifier chains method (ECC)~\cite{read2011classifier}. Note that the classification problem can be regarded as a regression problem\footnote{For a dataset for binary classification that each positive example has a label $+1$ and each negative example has a label $-1$, one can regard these labels as real numbers (i.e., $1.0$ for each of the positive examples and $-1.0$ for each of the negative examples). Then, this dataset can be used in a MTL method for regression to learn a regressor. After obtaining the regressor, for a test example $x$, if the predicted label of $x$ (by the regressor) is larger than $0$, one can regard $x$ as a positive example. On the other hand, if the predicted label of $x$ is smaller than $0$, then one can regard $x$ as a negative example.}.

The proposed methods, the baselines MTL-CLS and MTL-REG were implemented with Python 2.7. For MTL-REG, our implementations are based on the algorithms in~\cite{Liu} (for the $\ell_{2,1}$ norm) and~\cite{TraceNorm} (for the trace norm). According to Theorem 3 in~\cite{SVM-Perf}, the problem of MTL-CLS is equivalent to a special form of SMTL in (2) (with $\Delta(y^{(i)},y)=2\times t$, where $t$ represents the number of index $k$ that satisfies $y^{(i)}_k \neq y_k$). Hence, our implementation of MTL-CLS is based on the framework of Algorithm 1. For StructSVM, we use the open-source implementation of SVM-Perf~\cite{SVM-Perf}. All the experiments were conducted
on a Dell PowerEdge R320 server with 16G memory and 1.9Ghz E5-2420 CPU.

We report the experimental results on $9$ real-world datasets. The statistics of these datasets are summarized in Table \ref{datasets}. In the Emotions dataset, the labels are $6$ kinds of emotions, and the features are rhythmic and timbre extracted from music wave files. In the Yeast dataset, the labels are localization sites of protein, and the features are protein properties. In the Flags dataset, the labels are religions of countries and the features are extracted from flag images. In the Cal500 dataset, the labels are semantically meaning of popular songs and the features are extracted from audio data. In the Segmentation dataset, the labels are content of image region, and the features are pixels' properties of image regions. In the Optdigits dataset, the labels are handwritten digits $0$ to $9$, and the features are pixels. In the MediaMill dataset, the labels are semantic concepts of each video and the features are extracted from videos. In the TMC2007 dataset, the labels are the document topics, and the features are discrete attributes about terms. In the Scene dataset, the labels are scene types, and the features are spatial color moments in LUV space.
All of these datasets are normalized.
\begin{table}[htbp]
\scriptsize
\centering
\caption{Statistics of 9 datasets}
\begin{tabular}{|r|r|r|r|r|}
	\hline
	\textbf{} & Type & Features & Samples & Tasks \bigstrut\\
	\hline
	\textbf{Emotions} & \multicolumn{1}{l|}{music} & \multicolumn{1}{l|}{72} & \multicolumn{1}{l|}{593} & \multicolumn{1}{l|}{6} \bigstrut\\
	\hline
	\textbf{Yeast} & \multicolumn{1}{l|}{gene} & \multicolumn{1}{l|}{103} & \multicolumn{1}{l|}{2417} & \multicolumn{1}{l|}{14} \bigstrut\\
	\hline
	\textbf{Flags} & \multicolumn{1}{l|}{image} & \multicolumn{1}{l|}{19} & \multicolumn{1}{l|}{194} & \multicolumn{1}{l|}{7} \bigstrut\\
	\hline
	\textbf{Cal500} & \multicolumn{1}{l|}{songs } & \multicolumn{1}{l|}{68} & \multicolumn{1}{l|}{502} & \multicolumn{1}{l|}{174} \bigstrut\\
	\hline
	\textbf{Segmentation} & \multicolumn{1}{l|}{image} & \multicolumn{1}{l|}{19} & \multicolumn{1}{l|}{2310} & \multicolumn{1}{l|}{7} \bigstrut\\
	\hline
	\textbf{Optdigits} & \multicolumn{1}{l|}{image} & \multicolumn{1}{l|}{64} & \multicolumn{1}{l|}{5620} & \multicolumn{1}{l|}{10} \bigstrut\\
	\hline
	\textbf{MediaMill} & \multicolumn{1}{l|}{multimedia} & \multicolumn{1}{l|}{120} & \multicolumn{1}{l|}{10000} & \multicolumn{1}{l|}{12} \bigstrut\\
	\hline
	\textbf{TMC2007} & \multicolumn{1}{l|}{test} & \multicolumn{1}{l|}{500} & \multicolumn{1}{l|}{10000} & \multicolumn{1}{l|}{6} \bigstrut\\
	\hline
	\textbf{Scene} & \multicolumn{1}{l|}{image} & \multicolumn{1}{l|}{294} & \multicolumn{1}{l|}{2407} & \multicolumn{1}{l|}{6} \bigstrut\\
	\hline
\end{tabular}%
\label{datasets}%
\end{table}%
 \begin{table}[htbp]
 	\tiny
 	\centering
 	\caption{Comparison results on Cal500, Segmentation and Optdigits.}
 \begin{tabular}{|r|r|r|r|}
 	\hline
 	\multicolumn{1}{|c|}{\textbf{}} &
 	\multicolumn{1}{c|}{\textbf{MACRO }} &
 	\multicolumn{1}{c|}{\textbf{MICRO}} &
 	\multicolumn{1}{c|}{\textbf{Average}}
 	\bigstrut[t]\\
 	\multicolumn{1}{|c|}{\textbf{METHOD}} &
 	\multicolumn{1}{c|}{\boldmath{}\textbf{$F_1$}\unboldmath{}} &
 	\multicolumn{1}{c|}{\boldmath{}\textbf{$F_1$}\unboldmath{}} &
 	\multicolumn{1}{c|}{\textbf{AUC}}
 	\bigstrut[b]\\
 	\hline
 	\multicolumn{4}{|c|}{\textbf{Cal500}}
 	\bigstrut\\
 	\hline
 	\boldmath{}\textbf{SMTL($\ell_{2,1}$+AUC)}\unboldmath{} &
 	\boldmath{}\textbf{21.722$\pm$0.456}\unboldmath{} &
 	38.452$\pm$0.610 &
 	\boldmath{}\textbf{56.505$\pm$0.511}\unboldmath{}
 	\bigstrut[t]\\
 	\boldmath{}\textbf{SMTL($\ell_{2,1}$+$F_1$)}\unboldmath{} &
 	21.495$\pm$0.232 &
 	\boldmath{}\textbf{40.127$\pm$0.173}\unboldmath{} &
 	53.690$\pm$0.293
 	\\
 	MTL-CLS($\ell_{2,1}$) &
 	13.157$\pm$0.449 &
 	37.357$\pm$0.180 &
 	55.764$\pm$0.820
 	\\
 	MTL-REG($\ell_{2,1}$) &
 	12.500$\pm$0.129 &
 	36.438$\pm$0.176 &
 	52.964$\pm$0.758
 	\bigstrut[b]\\
 	\hline
 	\boldmath{}\textbf{SMTL($\ell_{1,1}$+AUC)}\unboldmath{} &
 	\boldmath{}\textbf{21.721$\pm$0.807}\unboldmath{} &
 	35.52$\pm$0.811  &
 	\boldmath{}\textbf{56.716$\pm$0.500}\unboldmath{}
 	\bigstrut[t]\\
 	\boldmath{}\textbf{SMTL($\ell_{1,1}$+$F_1$)}\unboldmath{} &
 	21.138$\pm$0.191 &
 	\boldmath{}\textbf{38.386$\pm$0.456}\unboldmath{} &
 	53.358$\pm$0.827
 	\\
 	MTL-CLS($\ell_{1,1}$) &
 	12.176$\pm$0.445 &
 	37.387$\pm$0.845 &
 	56.316$\pm$0.216
 	\\
 	MTL-REG($\ell_{1,1}$) &
 	12.447$\pm$0.297 &
 	36.66$\pm$0.638 &
 	53.628$\pm$0.264
 	\bigstrut[b]\\
 	\hline
 	\textbf{SMTL(TraceNorm+AUC)} &
 	\boldmath{}\textbf{21.772$\pm$0.545}\unboldmath{} &
 	35.204$\pm$0.585 &
 	\boldmath{}\textbf{56.798$\pm$0.358}\unboldmath{}
 	\bigstrut[t]\\
 	\boldmath{}\textbf{SMTL(TraceNorm+$F_1$)}\unboldmath{} &
 	21.768$\pm$0.333 &
 	\boldmath{}\textbf{38.559$\pm$0.394}\unboldmath{} &
 	54.987$\pm$0.823
 	\\
 	MTL-CLS(TraceNorm) &
 	12.884$\pm$0.353 &
 	37.402$\pm$0.501 &
 	55.635$\pm$0.511
 	\\
 	MTL-REG(TraceNorm) &
 	8.348$\pm$0.999 &
 	34.832$\pm$0.698 &
 	55.69$\pm$0.636
 	\bigstrut[b]\\
 	\hline
 	StructSVM &
 	20.864$\pm$1.150 &
 	35.408$\pm$1.150 &
 	51.427$\pm$0.841
 	\bigstrut[t]\\
 	RAkEL &
 	20.628$\pm$0.611 &
 	33.689$\pm$0.843 &
 	54.637$\pm$0.656
 	\\
 	MLCSSP &
 	21.677$\pm$0.514 &
 	27.093$\pm$0.537 &
 	52.69$\pm$0.983
 	\\
 	AdaBoostMH &
 	0.923$\pm$0.274 &
 	6.492$\pm$0.146 &
 	50.734$\pm$0.538
 	\\
 	HOMER &
 	13.850$\pm$0.163 &
 	30.332$\pm$1.313 &
 	52.461$\pm$0.937
 	\\
 	BR &
 	17.094$\pm$0.634 &
 	33.619$\pm$0.375 &
 	50.563$\pm$0.153
 	\\
 	LP &
 	15.257$\pm$0.428 &
 	32.978$\pm$0.668 &
 	52.117$\pm$0.685
 	\\
 	ECC &
 	9.600$\pm$0.666  &
 	34.789$\pm$0.482 &
 	52.117$\pm$0.625
 	\bigstrut[b]\\
 	\hline
 	\multicolumn{4}{|c|}{\textbf{Segmentation}}
 	\bigstrut\\
 	\hline
 	\boldmath{}\textbf{SMTL($\ell_{2,1}$+AUC)}\unboldmath{} &
 	72.832$\pm$1.567 &
 	68.445$\pm$1.543 &
 	\boldmath{}\textbf{97.195$\pm$0.4549}\unboldmath{}
 	\bigstrut[t]\\
 	\boldmath{}\textbf{SMTL($\ell_{2,1}$+$F_1$)}\unboldmath{} &
 	\boldmath{}\textbf{85.61$\pm$1.304}\unboldmath{} &
 	84.149$\pm$1.684 &
 	96.967$\pm$0.647
 	\\
 	MTL-CLS($\ell_{2,1}$) &
 	85.114$\pm$1.946 &
 	\boldmath{}\textbf{84.228$\pm$4.508}\unboldmath{} &
 	96.93$\pm$0.560
 	\\
 	MTL-REG($\ell_{2,1}$) &
 	75.547$\pm$1.215 &
 	81.702$\pm$2.456 &
 	96.757$\pm$0.645
 	\bigstrut[b]\\
 	\hline
 	\boldmath{}\textbf{SMTL($\ell_{1,1}$+AUC)}\unboldmath{} &
 	73.378$\pm$1.564 &
 	68.424$\pm$1.787 &
 	\boldmath{}\textbf{97.527$\pm$0.286}\unboldmath{}
 	\bigstrut[t]\\
 	\boldmath{}\textbf{SMTL($\ell_{1,1}$+$F_1$)}\unboldmath{} &
 	\boldmath{}\textbf{85.105$\pm$1.830}\unboldmath{} &
 	\boldmath{}\textbf{83.693$\pm$1.192}\unboldmath{} &
 	96.757$\pm$0.192
 	\\
 	MTL-CLS($\ell_{1,1}$) &
 	83.712$\pm$3.513 &
 	82.518$\pm$4.003 &
 	96.781$\pm$0.828
 	\\
 	MTL-REG($\ell_{1,1}$) &
 	76.253$\pm$2.564 &
 	82.606$\pm$0.156 &
 	96.798$\pm$0.231
 	\bigstrut[b]\\
 	\hline
 	\textbf{SMTL(TraceNorm+AUC)} &
 	72.265$\pm$1.453 &
 	67.655$\pm$1.978 &
 	\boldmath{}\textbf{97.134$\pm$0.457}\unboldmath{}
 	\bigstrut[t]\\
 	\boldmath{}\textbf{SMTL(TraceNorm+$F_1$)}\unboldmath{} &
 	\boldmath{}\textbf{85.356$\pm$1.092}\unboldmath{} &
 	\boldmath{}\textbf{83.462$\pm$1.805}\unboldmath{} &
 	96.863$\pm$0.322
 	\\
 	MTL-CLS(TraceNorm) &
 	82.703$\pm$3.865 &
 	82.150$\pm$5.439 &
 	96.705$\pm$0.612
 	\\
 	MTL-REG(TraceNorm) &
 	76.602$\pm$1.286 &
 	82.805$\pm$1.877 &
 	96.698$\pm$0.147
 	\bigstrut[b]\\
 	\hline
 	StructSVM &
 	44.632$\pm$1.828 &
 	53.992$\pm$1.828 &
 	89.355$\pm$0.311
 	\bigstrut[t]\\
 	RAkEL &
 	75.592$\pm$0.243 &
 	70.980$\pm$0.398 &
 	91.333$\pm$0.082
 	\\
 	MLCSSP &
 	79.821$\pm$8.533 &
 	78.923$\pm$14.036 &
 	93.810$\pm$0.329
 	\\
 	AdaBoostMH &
 	75.633$\pm$0.209 &
 	71.018$\pm$0.376 &
 	96.148$\pm$0.089
 	\\
 	HOMER &
 	72.920$\pm$2.505 &
 	69.969$\pm$1.651 &
 	91.225$\pm$1.543
 	\\
 	BR &
 	84.236$\pm$0.638 &
 	78.796$\pm$0.708 &
 	96.870$\pm$0.194
 	\\
 	LP &
 	84.394$\pm$0.603 &
 	83.411$\pm$0.615 &
 	96.240$\pm$0.124
 	\\
 	ECC &
 	84.183$\pm$0.550 &
 	82.942$\pm$0.542 &
 	96.782$\pm$0.269
 	\bigstrut[b]\\
 	\hline
 	\multicolumn{4}{|c|}{\textbf{Optdigits}}
 	\bigstrut\\
 	\hline
 	\boldmath{}\textbf{SMTL($\ell_{2,1}$+AUC)}\unboldmath{} &
 	92.722$\pm$0.595 &
 	92.734$\pm$0.712 &
 	\boldmath{}\textbf{99.657$\pm$0.0528}\unboldmath{}
 	\bigstrut[t]\\
 	\boldmath{}\textbf{SMTL($\ell_{2,1}$+$F_1$)}\unboldmath{} &
 	\boldmath{}\textbf{93.963$\pm$0.164}\unboldmath{} &
 	\boldmath{}\textbf{93.964$\pm$0.235}\unboldmath{} &
 	99.589$\pm$0.054
 	\\
 	MTL-CLS($\ell_{2,1}$) &
 	93.701$\pm$0.403 &
 	92.773$\pm$0.440 &
 	99.206$\pm$0.044
 	\\
 	MTL-REG($\ell_{2,1}$) &
 	88.901$\pm$0.306 &
 	89.268$\pm$0.875 &
 	99.32$\pm$0.089
 	\bigstrut[b]\\
 	\hline
 	\boldmath{}\textbf{SMTL($\ell_{1,1}$+AUC)}\unboldmath{} &
 	92.526$\pm$0.624 &
 	92.213$\pm$0.670 &
 	\boldmath{}\textbf{99.653$\pm$0.078}\unboldmath{}
 	\bigstrut[t]\\
 	\boldmath{}\textbf{SMTL($\ell_{1,1}$+$F_1$)}\unboldmath{} &
 	\boldmath{}\textbf{93.692$\pm$0.508}\unboldmath{} &
 	\boldmath{}\textbf{94.626$\pm$0.520}\unboldmath{} &
 	99.554$\pm$0.047
 	\\
 	MTL-CLS($\ell_{1,1}$) &
 	92.961$\pm$0.608 &
 	94.009$\pm$0.356 &
 	98.658$\pm$0.067
 	\\
 	MTL-REG($\ell_{1,1}$) &
 	88.762$\pm$0.845 &
 	89.203$\pm$0.865 &
 	99.269$\pm$0.045
 	\bigstrut[b]\\
 	\hline
 	\textbf{SMTL(TraceNorm+AUC)} &
 	92.862$\pm$0.543 &
 	92.802$\pm$0.944 &
 	\boldmath{}\textbf{99.654$\pm$0.036}\unboldmath{}
 	\bigstrut[t]\\
 	\boldmath{}\textbf{SMTL(TraceNorm+$F_1$)}\unboldmath{} &
 	\boldmath{}\textbf{94.206$\pm$0.202}\unboldmath{} &
 	\boldmath{}\textbf{94.139$\pm$0.266}\unboldmath{} &
 	99.566$\pm$0.027
 	\\
 	MTL-CLS(TraceNorm) &
 	93.701$\pm$0.435 &
 	93.773$\pm$0.267 &
 	99.182$\pm$0.065
 	\\
 	MTL-REG(TraceNorm) &
 	88.777$\pm$0.765 &
 	89.173$\pm$0.946 &
 	99.293$\pm$0.048
 	\bigstrut[b]\\
 	\hline
 	StructSVM &
 	36.276$\pm$0.905 &
 	38.289$\pm$2.218 &
 	98.400$\pm$0.366
 	\bigstrut[t]\\
 	RAkEL &
 	82.450$\pm$0.168 &
 	80.967$\pm$0.311 &
 	94.543$\pm$0.070
 	\\
 	MLCSSP &
 	75.191$\pm$2.245 &
 	82.129$\pm$3.195 &
 	88.879$\pm$0.195
 	\\
 	AdaBoostMH &
 	93.083$\pm$0.695 &
 	93.108$\pm$0.669 &
 	98.594$\pm$0.119
 	\\
 	HOMER &
 	74.869$\pm$4.151 &
 	75.713$\pm$3.663 &
 	93.391$\pm$0.964
 	\\
 	BR &
 	92.625$\pm$0.348 &
 	92.714$\pm$0.383 &
 	99.370$\pm$0.122
 	\\
 	LP &
 	88.875$\pm$0.212 &
 	88.915$\pm$0.269 &
 	94.941$\pm$0.329
 	\\
 	ECC &
 	93.043$\pm$0.206 &
 	94.019$\pm$0.213 &
 	99.019$\pm$0.156
 	\bigstrut[b]\\
 	\hline
 \end{tabular}%

 	\label{first_experiment1}%
 \end{table}%

 \begin{table}[htbp]
 	\tiny
 	\centering
 	\caption{Comparison results on Scene, MediaMill and TMC2007.}
 \begin{tabular}{|r|r|r|r|}
 	\hline
 	\multicolumn{1}{|c|}{\textbf{}} &
 	\multicolumn{1}{c|}{\textbf{MACRO }} &
 	\multicolumn{1}{c|}{\textbf{MICRO}} &
 	\multicolumn{1}{c|}{\textbf{Average}}
 	\bigstrut[t]\\
 	\multicolumn{1}{|c|}{\textbf{METHOD}} &
 	\multicolumn{1}{c|}{\boldmath{}\textbf{$F_1$}\unboldmath{}} &
 	\multicolumn{1}{c|}{\boldmath{}\textbf{$F_1$}\unboldmath{}} &
 	\multicolumn{1}{c|}{\textbf{AUC}}
 	\bigstrut[b]\\
 	\hline
 	\multicolumn{4}{|c|}{\textbf{Scene}}
 	\bigstrut\\
 	\hline
 	\boldmath{}\textbf{SMTL($\ell_{2,1}$+AUC)}\unboldmath{} &
 	54.013$\pm$1.124 &
 	54.746$\pm$1.231 &
 	\boldmath{}\textbf{89.99$\pm$0.820}\unboldmath{}
 	\bigstrut[t]\\
 	\boldmath{}\textbf{SMTL($\ell_{2,1}$+$F_1$)}\unboldmath{} &
 	\boldmath{}\textbf{55.787$\pm$0.756}\unboldmath{} &
 	\boldmath{}\textbf{56.434$\pm$0.567}\unboldmath{} &
 	87.652$\pm$0.280
 	\\
 	MTL-CLS($\ell_{2,1}$) &
 	54.722$\pm$1.590 &
 	54.508$\pm$1.176 &
 	86.738$\pm$1.102
 	\\
 	MTL-REG($\ell_{2,1}$) &
 	51.157$\pm$0.343 &
 	52.810$\pm$0.345 &
 	85.194$\pm$0.712
 	\bigstrut[b]\\
 	\hline
 	\boldmath{}\textbf{SMTL($\ell_{1,1}$+AUC)}\unboldmath{} &
 	54.296$\pm$0.977 &
 	54.333$\pm$0.025 &
 	\boldmath{}\textbf{88.358$\pm$0.467}\unboldmath{}
 	\bigstrut[t]\\
 	\boldmath{}\textbf{SMTL($\ell_{1,1}$+$F_1$)}\unboldmath{} &
 	\boldmath{}\textbf{55.501$\pm$1.92}\unboldmath{} &
 	\boldmath{}\textbf{56.007$\pm$2.34}\unboldmath{} &
 	87.364$\pm$1.801
 	\\
 	MTL-CLS($\ell_{1,1}$) &
 	54.387$\pm$0.730 &
 	54.805$\pm$1.488 &
 	85.952$\pm$1.116
 	\\
 	MTL-REG($\ell_{1,1}$) &
 	50.748$\pm$0.546 &
 	51.280$\pm$0.619 &
 	85.032$\pm$0.779
 	\bigstrut[b]\\
 	\hline
 	\textbf{SMTL(TraceNorm+AUC)} &
 	54.227$\pm$0.660 &
 	55.384$\pm$0.804 &
 	\boldmath{}\textbf{88.421$\pm$1.103}\unboldmath{}
 	\bigstrut[t]\\
 	\boldmath{}\textbf{SMTL(TraceNorm+$F_1$)}\unboldmath{} &
 	\boldmath{}\textbf{55.396$\pm$1.089}\unboldmath{} &
 	\boldmath{}\textbf{56.304$\pm$1.119}\unboldmath{} &
 	87.071$\pm$0.682
 	\\
 	MTL-CLS(TraceNorm) &
 	55.104$\pm$0.298 &
 	55.481$\pm$0.506 &
 	86.205$\pm$0.471
 	\\
 	MTL-REG(TraceNorm) &
 	50.832$\pm$0.226 &
 	51.236$\pm$0.264 &
 	85.275$\pm$0.852
 	\bigstrut[b]\\
 	\hline
 	StructSVM &
 	49.826$\pm$0.815 &
 	49.951$\pm$0.755 &
 	82.375$\pm$0.393
 	\bigstrut[t]\\
 	RAkEL &
 	54.592$\pm$0.613 &
 	55.719$\pm$0.565 &
 	78.981$\pm$0.535
 	\\
 	MLCSSP &
 	42.764$\pm$0.080 &
 	47.178$\pm$0.181 &
 	65.830$\pm$2.240
 	\\
 	AdaBoostMH &
 	36.506$\pm$0.404 &
 	40.681$\pm$0.449 &
 	87.617$\pm$0.470
 	\\
 	HOMER &
 	60.980$\pm$2.470 &
 	58.251$\pm$2.592 &
 	80.744$\pm$0.360
 	\\
 	BR &
 	54.579$\pm$1.813 &
 	55.019$\pm$1.843 &
 	82.888$\pm$1.164
 	\\
 	LP &
 	54.902$\pm$1.503 &
 	55.818$\pm$1.595 &
 	75.900$\pm$1.362
 	\\
 	ECC &
 	55.347$\pm$0.893 &
 	55.831$\pm$0.881 &
 	88.153$\pm$0.298
 	\bigstrut[b]\\
 	\hline
 	\multicolumn{4}{|c|}{\textbf{MediaMill}}
 	\bigstrut\\
 	\hline
 	\boldmath{}\textbf{SMTL($\ell_{2,1}$+AUC)}\unboldmath{} &
 	18.030$\pm$0.294 &
 	22.058$\pm$0.257 &
 	66.068$\pm$0.426
 	\bigstrut[t]\\
 	\boldmath{}\textbf{SMTL($\ell_{2,1}$+$F_1$)}\unboldmath{} &
 	\boldmath{}\textbf{22.851$\pm$5.093}\unboldmath{} &
 	\boldmath{}\textbf{56.424$\pm$2.761}\unboldmath{} &
 	\boldmath{}\textbf{78.705$\pm$2.280}\unboldmath{}
 	\\
 	MTL-CLS($\ell_{2,1}$) &
 	10.613$\pm$1.733 &
 	55.441$\pm$3.647 &
 	76.216$\pm$2.474
 	\\
 	MTL-REG($\ell_{2,1}$) &
 	6.366$\pm$0.065 &
 	55.515$\pm$0.465 &
 	53.867$\pm$0.496
 	\bigstrut[b]\\
 	\hline
 	\boldmath{}\textbf{SMTL($\ell_{1,1}$+AUC)}\unboldmath{} &
 	18.012$\pm$0.286 &
 	22.232$\pm$0.211 &
 	65.405$\pm$0.503
 	\bigstrut[t]\\
 	\boldmath{}\textbf{SMTL($\ell_{1,1}$+$F_1$)}\unboldmath{} &
 	\boldmath{}\textbf{22.386$\pm$5.326}\unboldmath{} &
 	\boldmath{}\textbf{56.169$\pm$2.436}\unboldmath{} &
 	\boldmath{}\textbf{78.907$\pm$1.854}\unboldmath{}
 	\\
 	MTL-CLS($\ell_{1,1}$) &
 	8.542$\pm$1.672 &
 	55.838$\pm$2.229 &
 	74.037$\pm$1.219
 	\\
 	MTL-REG($\ell_{1,1}$) &
 	6.393$\pm$0.033 &
 	55.687$\pm$0.439 &
 	53.036$\pm$0.181
 	\bigstrut[b]\\
 	\hline
 	\textbf{SMTL(TraceNorm+AUC)} &
 	18.201$\pm$0.221 &
 	22.684$\pm$0.354 &
 	66.847$\pm$1.015
 	\bigstrut[t]\\
 	\boldmath{}\textbf{SMTL(TraceNorm+$F_1$)}\unboldmath{} &
 	\boldmath{}\textbf{27.973$\pm$3.006}\unboldmath{} &
 	\boldmath{}\textbf{56.031$\pm$4.924}\unboldmath{} &
 	\boldmath{}\textbf{79.730$\pm$1.850}\unboldmath{}
 	\\
 	MTL-CLS(TraceNorm) &
 	15.800$\pm$0.589 &
 	50.098$\pm$5.569 &
 	75.968$\pm$2.144
 	\\
 	MTL-REG(TraceNorm) &
 	6.380$\pm$0.045 &
 	55.333$\pm$0.425 &
 	53.825$\pm$0.493
 	\bigstrut[b]\\
 	\hline
 	StructSVM &
 	17.847$\pm$0.318 &
 	22.030$\pm$0.284 &
 	64.761$\pm$0.487
 	\bigstrut[t]\\
 	RAkEL &
 	19.874$\pm$0.156 &
 	26.686$\pm$0.189 &
 	63.241$\pm$0.398
 	\\
 	MLCSSP &
 	15.129$\pm$0.633 &
 	20.124$\pm$0.723 &
 	52.473$\pm$1.884
 	\\
 	AdaBoostMH &
 	17.939$\pm$0.469 &
 	41.991$\pm$0.425 &
 	61.914$\pm$0.167
 	\\
 	HOMER &
 	17.939$\pm$0.469 &
 	41.991$\pm$0.425 &
 	61.914$\pm$0.167
 	\\
 	BR &
 	19.769$\pm$0.196 &
 	26.515$\pm$0.166 &
 	69.032$\pm$0.854
 	\\
 	LP &
 	24.135$\pm$0.959 &
 	50.170$\pm$0.402 &
 	60.597$\pm$0.502
 	\\
 	ECC &
 	24.879$\pm$0.590 &
 	56.214$\pm$0.363 &
 	78.067$\pm$0.705
 	\bigstrut[b]\\
 	\hline
 	\multicolumn{4}{|c|}{\textbf{TMC2007}}
 	\bigstrut\\
 	\hline
 	\boldmath{}\textbf{SMTL($\ell_{2,1}$+AUC)}\unboldmath{} &
 	59.432$\pm$0.581 &
 	68.02$\pm$1.042 &
 	90.138$\pm$0.17
 	\bigstrut[t]\\
 	\boldmath{}\textbf{SMTL($\ell_{2,1}$+$F_1$)}\unboldmath{} &
 	\boldmath{}\textbf{64.321$\pm$0.955}\unboldmath{} &
 	\boldmath{}\textbf{74.159$\pm$0.255}\unboldmath{} &
 	\boldmath{}\textbf{90.561$\pm$0.669}\unboldmath{}
 	\\
 	MTL-CLS($\ell_{2,1}$) &
 	60.517$\pm$1.363 &
 	71.284$\pm$0.387 &
 	88.382$\pm$0.398
 	\\
 	MTL-REG($\ell_{2,1}$) &
 	37.106$\pm$0.416 &
 	70.181$\pm$0.221 &
 	85.218$\pm$0.529
 	\bigstrut[b]\\
 	\hline
 	\boldmath{}\textbf{SMTL($\ell_{1,1}$+AUC)}\unboldmath{} &
 	60.249$\pm$0.147 &
 	67.654$\pm$0.234 &
 	\boldmath{}\textbf{90.441$\pm$0.077}\unboldmath{}
 	\bigstrut[t]\\
 	\boldmath{}\textbf{SMTL($\ell_{1,1}$+$F_1$)}\unboldmath{} &
 	\boldmath{}\textbf{65.436$\pm$1.239}\unboldmath{} &
 	\boldmath{}\textbf{73.984$\pm$0.533}\unboldmath{} &
 	90.238$\pm$0.732
 	\\
 	MTL-CLS($\ell_{1,1}$) &
 	62.919$\pm$0.802 &
 	72.745$\pm$0.464 &
 	89.074$\pm$0.59
 	\\
 	MTL-REG($\ell_{1,1}$) &
 	37.709$\pm$0.32 &
 	70.431$\pm$0.414 &
 	86.612$\pm$0.592
 	\bigstrut[b]\\
 	\hline
 	\textbf{SMTL(TraceNorm+AUC)} &
 	58.595$\pm$0.148 &
 	68.056$\pm$0.45 &
 	88.325$\pm$0.182
 	\bigstrut[t]\\
 	\boldmath{}\textbf{SMTL(TraceNorm+$F_1$)}\unboldmath{} &
 	\boldmath{}\textbf{61.867$\pm$1.014}\unboldmath{} &
 	\boldmath{}\textbf{72.588$\pm$0.350}\unboldmath{} &
 	\boldmath{}\textbf{89.328$\pm$0.815}\unboldmath{}
 	\\
 	MTL-CLS(TraceNorm) &
 	59.752$\pm$0.951 &
 	71.863$\pm$0.628 &
 	87.933$\pm$0.428
 	\\
 	MTL-REG(TraceNorm) &
 	36.64$\pm$0.314 &
 	70.118$\pm$0.437 &
 	84.54$\pm$0.743
 	\bigstrut[b]\\
 	\hline
 	StructSVM &
 	37.19$\pm$0.652 &
 	45.027$\pm$0.601 &
 	88.072$\pm$0.289
 	\bigstrut[t]\\
 	RAkEL &
 	57.331$\pm$0.592 &
 	69.813$\pm$0.179 &
 	81.994$\pm$0.134
 	\\
 	MLCSSP &
 	56.717$\pm$0.790 &
 	60.417$\pm$1.665 &
 	75.246$\pm$1.093
 	\\
 	AdaBoostMH &
 	15.170$\pm$1.893 &
 	56.004$\pm$1.103 &
 	61.466$\pm$0.206
 	\\
 	HOMER &
 	61.144$\pm$0.238 &
 	71.429$\pm$0.104 &
 	84.998$\pm$0.589
 	\\
 	BR &
 	51.939$\pm$1.225 &
 	67.873$\pm$0.374 &
 	84.616$\pm$0.528
 	\\
 	LP &
 	52.683$\pm$0.832 &
 	62.672$\pm$0.526 &
 	73.063$\pm$0.637
 	\\
 	ECC &
 	58.368$\pm$0.714 &
 	68.223$\pm$0.096 &
 	86.287$\pm$0.664
 	\bigstrut[b]\\
 	\hline
 \end{tabular}%

 	\label{first_experiment2}%
 \end{table}%

 \begin{table}[htbp]
 	\tiny
 	\centering
 	\caption{Comparison results Emotions, Yeast and Flags.}
 	
	 \begin{tabular}{|r|r|r|r|}
	 	\hline
	 	\multicolumn{1}{|c|}{\textbf{}} &
	 	\multicolumn{1}{c|}{\textbf{MACRO }} &
	 	\multicolumn{1}{c|}{\textbf{MICRO}} &
	 	\multicolumn{1}{c|}{\textbf{Average}}
	 	\bigstrut[t]\\
	 	\multicolumn{1}{|c|}{\textbf{METHOD}} &
	 	\multicolumn{1}{c|}{\boldmath{}\textbf{$F_1$}\unboldmath{}} &
	 	\multicolumn{1}{c|}{\boldmath{}\textbf{$F_1$}\unboldmath{}} &
	 	\multicolumn{1}{c|}{\textbf{AUC}}
	 	\bigstrut[b]\\
	 	\hline
	 	\multicolumn{4}{|c|}{\textbf{Emotions}}
	 	\bigstrut\\
	 	\hline
	 	\boldmath{}\textbf{SMTL($\ell_{2,1}$+AUC)}\unboldmath{} &
	 	65.498$\pm$2.047 &
	 	\boldmath{}\textbf{67.067$\pm$1.956}\unboldmath{} &
	 	\boldmath{}\textbf{83.378$\pm$0.466}\unboldmath{}
	 	\bigstrut[t]\\
	 	\boldmath{}\textbf{SMTL($\ell_{2,1}$+$F_1$)}\unboldmath{} &
	 	\boldmath{}\textbf{66.244$\pm$1.584}\unboldmath{} &
	 	66.358$\pm$1.255 &
	 	81.986$\pm$0.495
	 	\\
	 	MTL-CLS($\ell_{2,1}$) &
	 	63.343$\pm$1.688 &
	 	65.684$\pm$1.327 &
	 	80.065$\pm$0.490
	 	\\
	 	MTL-REG($\ell_{2,1}$) &
	 	62.621$\pm$1.543 &
	 	63.701$\pm$1.054 &
	 	81.32$\pm$0.396
	 	\bigstrut[b]\\
	 	\hline
	 	\boldmath{}\textbf{SMTL($\ell_{1,1}$+AUC)}\unboldmath{} &
	 	65.622$\pm$1.984 &
	 	67.143$\pm$1.629 &
	 	\boldmath{}\textbf{83.358$\pm$0.345}\unboldmath{}
	 	\bigstrut[t]\\
	 	\boldmath{}\textbf{SMTL($\ell_{1,1}$+$F_1$)}\unboldmath{} &
	 	\boldmath{}\textbf{67.696$\pm$0.348}\unboldmath{} &
	 	\boldmath{}\textbf{67.923$\pm$0.578}\unboldmath{} &
	 	83.106$\pm$0.596
	 	\\
	 	MTL-CLS($\ell_{1,1}$) &
	 	64.969$\pm$0.822 &
	 	66.584$\pm$1.049 &
	 	80.03$\pm$0.574
	 	\\
	 	MTL-REG($\ell_{1,1}$) &
	 	62.976$\pm$0.547 &
	 	64.404$\pm$1.535 &
	 	81.811$\pm$0.587
	 	\bigstrut[b]\\
	 	\hline
	 	\textbf{SMTL(TraceNorm+AUC)} &
	 	65.902$\pm$1.904 &
	 	67.405$\pm$1.848 &
	 	\boldmath{}\textbf{83.362$\pm$0.618}\unboldmath{}
	 	\bigstrut[t]\\
	 	\boldmath{}\textbf{SMTL(TraceNorm+$F_1$)}\unboldmath{} &
	 	\boldmath{}\textbf{67.600$\pm$0.574}\unboldmath{} &
	 	\boldmath{}\textbf{67.858$\pm$0.984}\unboldmath{} &
	 	83.000$\pm$0.236
	 	\\
	 	MTL-CLS(TraceNorm) &
	 	63.805$\pm$2.339 &
	 	66.602$\pm$2.063 &
	 	80.485$\pm$0.597
	 	\\
	 	MTL-REG(TraceNorm) &
	 	63.243$\pm$1.574 &
	 	64.869$\pm$2.574 &
	 	82.834$\pm$0.266
	 	\bigstrut[b]\\
	 	\hline
	 	StructSVM &
	 	46.367$\pm$5.531 &
	 	49.902$\pm$19.032 &
	 	62.908$\pm$4.361
	 	\bigstrut[t]\\
	 	RAkEL &
	 	64.998$\pm$1.387 &
	 	65.835$\pm$1.136 &
	 	75.206$\pm$0.875
	 	\\
	 	MLCSSP &
	 	62.980$\pm$2.780 &
	 	63.593$\pm$2.603 &
	 	76.054$\pm$2.495
	 	\\
	 	AdaBoostMH &
	 	4.291$\pm$1.429 &
	 	7.577$\pm$2.627 &
	 	55.111$\pm$0.328
	 	\\
	 	HOMER &
	 	59.039$\pm$2.431 &
	 	61.830$\pm$1.642 &
	 	71.212$\pm$1.167
	 	\\
	 	BR &
	 	61.358$\pm$2.578 &
	 	62.635$\pm$2.332 &
	 	79.146$\pm$1.250
	 	\\
	 	LP &
	 	53.384$\pm$1.858 &
	 	54.618$\pm$1.543 &
	 	68.506$\pm$0.652
	 	\\
	 	ECC &
	 	62.694$\pm$1.645 &
	 	64.138$\pm$1.216 &
	 	82.589$\pm$1.131
	 	\bigstrut[b]\\
	 	\hline
	 	\multicolumn{4}{|c|}{\textbf{Yeast}}
	 	\bigstrut\\
	 	\hline
	 	\boldmath{}\textbf{SMTL($\ell_{2,1}$+AUC)}\unboldmath{} &
	 	43.593$\pm$1.120 &
	 	46.261$\pm$0.872 &
	 	\boldmath{}\textbf{63.018$\pm$1.504}\unboldmath{}
	 	\bigstrut[t]\\
	 	\boldmath{}\textbf{SMTL($\ell_{2,1}$+$F_1$)}\unboldmath{} &
	 	\boldmath{}\textbf{44.353$\pm$1.080}\unboldmath{} &
	 	\boldmath{}\textbf{55.451$\pm$0.457}\unboldmath{} &
	 	61.285$\pm$1.246
	 	\\
	 	MTL-CLS($\ell_{2,1}$) &
	 	36.308$\pm$0.974 &
	 	43.908$\pm$0.499 &
	 	56.686$\pm$0.539
	 	\\
	 	MTL-REG($\ell_{2,1}$) &
	 	28.187$\pm$1.544 &
	 	47.029$\pm$0.645 &
	 	62.757$\pm$1.745
	 	\bigstrut[b]\\
	 	\hline
	 	\boldmath{}\textbf{SMTL($\ell_{1,1}$+AUC)}\unboldmath{} &
	 	43.132$\pm$1.349 &
	 	45.729$\pm$1.643 &
	 	\boldmath{}\textbf{62.626$\pm$1.709}\unboldmath{}
	 	\bigstrut[t]\\
	 	\boldmath{}\textbf{SMTL($\ell_{1,1}$+$F_1$)}\unboldmath{} &
	 	\boldmath{}\textbf{44.647$\pm$1.058}\unboldmath{} &
	 	\boldmath{}\textbf{54.971$\pm$1.187}\unboldmath{} &
	 	61.569$\pm$1.945
	 	\\
	 	MTL-CLS($\ell_{1,1}$) &
	 	36.89$\pm$0.699 &
	 	44.620$\pm$0.553 &
	 	58.221$\pm$0.424
	 	\\
	 	MTL-REG($\ell_{1,1}$) &
	 	33.720$\pm$1.634 &
	 	54.682$\pm$1.846 &
	 	50.050$\pm$1.563
	 	\bigstrut[b]\\
	 	\hline
	 	\textbf{SMTL(TraceNorm+AUC)} &
	 	43.58$\pm$1.046 &
	 	46.395$\pm$1.067 &
	 	\boldmath{}\textbf{63.058$\pm$0.634}\unboldmath{}
	 	\bigstrut[t]\\
	 	\boldmath{}\textbf{SMTL(TraceNorm+$F_1$)}\unboldmath{} &
	 	\boldmath{}\textbf{44.972$\pm$0.765}\unboldmath{} &
	 	\boldmath{}\textbf{50.471$\pm$0.968}\unboldmath{} &
	 	61.819$\pm$0.395
	 	\\
	 	MTL-CLS(TraceNorm) &
	 	42.275$\pm$1.006 &
	 	44.542$\pm$0.460 &
	 	61.528$\pm$0.590
	 	\\
	 	MTL-REG(TraceNorm) &
	 	28.178$\pm$1.043 &
	 	47.046$\pm$0.126 &
	 	62.920$\pm$0.326
	 	\bigstrut[b]\\
	 	\hline
	 	StructSVM &
	 	42.669$\pm$ 2.48 &
	 	46.298$\pm$2.048 &
	 	61.894$\pm$2.488
	 	\bigstrut[t]\\
	 	RAkEL &
	 	44.101$\pm$0.389 &
	 	46.086$\pm$0.450 &
	 	61.971$\pm$0.753
	 	\\
	 	MLCSSP &
	 	41.511$\pm$0.837 &
	 	46.200$\pm$1.272 &
	 	50.756$\pm$0.451
	 	\\
	 	AdaBoostMH &
	 	12.255$\pm$0.041 &
	 	48.144$\pm$0.315 &
	 	50.805$\pm$0.050
	 	\\
	 	HOMER &
	 	40.054$\pm$1.063 &
	 	53.745$\pm$0.867 &
	 	62.311$\pm$1.265
	 	\\
	 	BR &
	 	39.209$\pm$0.891 &
	 	54.153$\pm$0.543 &
	 	62.375$\pm$0.408
	 	\\
	 	LP &
	 	37.029$\pm$0.584 &
	 	53.059$\pm$0.514 &
	 	56.616$\pm$1.394
	 	\\
	 	ECC &
	 	37.523$\pm$0.310 &
	 	54.632$\pm$0.325 &
	 	62.105$\pm$0.627
	 	\bigstrut[b]\\
	 	\hline
	 	\multicolumn{4}{|c|}{\textbf{Flags}}
	 	\bigstrut\\
	 	\hline
	 	\boldmath{}\textbf{SMTL($\ell_{2,1}$+AUC)}\unboldmath{} &
	 	60.473$\pm$1.951 &
	 	61.666$\pm$2.226 &
	 	73.875$\pm$2.563
	 	\bigstrut[t]\\
	 	\boldmath{}\textbf{SMTL($\ell_{2,1}$+$F_1$)}\unboldmath{} &
	 	\boldmath{}\textbf{70.279$\pm$1.744}\unboldmath{} &
	 	\boldmath{}\textbf{75.047$\pm$0.945}\unboldmath{} &
	 	\boldmath{}\textbf{75.000$\pm$0.745}\unboldmath{}
	 	\\
	 	MTL-CLS($\ell_{2,1}$) &
	 	65.233$\pm$1.930 &
	 	71.709$\pm$0.955 &
	 	72.928$\pm$1.479
	 	\\
	 	MTL-REG($\ell_{2,1}$) &
	 	66.073$\pm$0.276 &
	 	73.005$\pm$1.307 &
	 	71.429$\pm$1.105
	 	\bigstrut[b]\\
	 	\hline
	 	\boldmath{}\textbf{SMTL($\ell_{1,1}$+AUC)}\unboldmath{} &
	 	60.187$\pm$1.971 &
	 	61.618$\pm$1.714 &
	 	74.136$\pm$2.805
	 	\bigstrut[t]\\
	 	\boldmath{}\textbf{SMTL($\ell_{1,1}$+$F_1$)}\unboldmath{} &
	 	\boldmath{}\textbf{69.122$\pm$1.975}\unboldmath{} &
	 	\boldmath{}\textbf{74.259$\pm$1.378}\unboldmath{} &
	 	\boldmath{}\textbf{74.168$\pm$1.513}\unboldmath{}
	 	\\
	 	MTL-CLS($\ell_{1,1}$) &
	 	65.532$\pm$1.210 &
	 	72.666$\pm$1.752 &
	 	72.725$\pm$0.497
	 	\\
	 	MTL-REG($\ell_{1,1}$) &
	 	65.256$\pm$0.739 &
	 	72.246$\pm$0.928 &
	 	71.299$\pm$0.998
	 	\bigstrut[b]\\
	 	\hline
	 	\textbf{SMTL(TraceNorm+AUC)} &
	 	61.435$\pm$1.616 &
	 	62.84$\pm$1.481 &
	 	\boldmath{}\textbf{74.367$\pm$2.373}\unboldmath{}
	 	\bigstrut[t]\\
	 	\boldmath{}\textbf{SMTL(TraceNorm+$F_1$)}\unboldmath{} &
	 	\boldmath{}\textbf{68.704$\pm$1.650}\unboldmath{} &
	 	\boldmath{}\textbf{73.132$\pm$1.891}\unboldmath{} &
	 	73.145$\pm$1.973
	 	\\
	 	MTL-CLS(TraceNorm) &
	 	65.236$\pm$3.507 &
	 	72.688$\pm$2.156 &
	 	73.307$\pm$2.155
	 	\\
	 	MTL-REG(TraceNorm) &
	 	65.257$\pm$2.647 &
	 	72.437$\pm$1.918 &
	 	71.495$\pm$0.783
	 	\bigstrut[b]\\
	 	\hline
	 	StructSVM &
	 	55.683$\pm$5.777 &
	 	51.957$\pm$2.048 &
	 	72.178$\pm$3.604
	 	\bigstrut[t]\\
	 	RAkEL &
	 	60.696$\pm$5.216 &
	 	64.749$\pm$4.688 &
	 	61.260$\pm$3.805
	 	\\
	 	MLCSSP &
	 	59.629$\pm$1.619 &
	 	63.215$\pm$1.326 &
	 	55.865$\pm$1.909
	 	\\
	 	AdaBoostMH &
	 	56.457$\pm$4.288 &
	 	71.268$\pm$1.400 &
	 	69.329$\pm$2.043
	 	\\
	 	HOMER &
	 	59.018$\pm$1.269 &
	 	63.855$\pm$2.259 &
	 	64.826$\pm$0.569
	 	\\
	 	BR &
	 	59.421$\pm$2.163 &
	 	67.287$\pm$1.876 &
	 	66.823$\pm$2.860
	 	\\
	 	LP &
	 	61.801$\pm$3.822 &
	 	69.132$\pm$3.200 &
	 	60.540$\pm$4.149
	 	\\
	 	ECC &
	 	64.936$\pm$3.023 &
	 	72.715$\pm$1.675 &
	 	73.913$\pm$2.339
	 	\bigstrut[b]\\
	 	\hline
	 \end{tabular}%

 	\label{first_experiment3}%
 \end{table}%

Following the settings in~\cite{Chen10}, to evaluate the performance, we use AUC, Macro F1-score, and Micro F1-score as the evaluation metrics (the details about the computation of AUC and $F_1$\footnote{In MTL, the Macro $F_1$ is calculated by firstly calculating the $F_1$ score of each individual task, and then average these $F_1$ scores over all tasks. The Micro $F_1$ in MTL is calculated by $\frac{2\times P \times R}{P+R}$, where
	
    \begin{displaymath}
    P = \frac{\sum_{i=1}^m\sum_{k=1}^{n_i}I({\mathbf{y}_k^{(i)}}=1\ \textit{and}\ (\mathbf{y}_j)_k=1)}{\sum_{i=1}^m\sum_{k=1}^{n_i}I({\mathbf{y}_k^{(i)}}=1)},
    \end{displaymath}
    \begin{displaymath}
     R = \frac{\sum_{i=1}^m\sum_{k=1}^{n_i}I({\mathbf{y}_k^{(i)}}=1 \ \textit{and}\ (\mathbf{y}_j)_k=1)}{\sum_{i=1}^m\sum_{k=1}^{n_i}I({(\mathbf{y}_j)_k}=1)}.
    \end{displaymath}} can be found in Section 4).

For each dataset, we firstly generate 10 $60\%$:$40\%$ partitions. In each partition, the ``$60\%$'' part is used as the training set and the ``$40\%$'' part is used as the test set. Then, we run each of the methods (the baselines and the proposed methods) on these 10 partitions, and reported the averaged results on these $10$ trials. Note that, for a fair comparison, in a dataset, each method uses the same ten partitions to produce its results. After the training set is determined, we conduct 10-fold cross validation on the training set to choose the trade-off parameter $\lambda$ within $\{ {10^{ - 3}} \times i\} _{i = 1}^{10} \cup \{ {10^{ - 2}} \times i\} _{i = 1}^{10} \cup \{ {10^{ - 1}} \times i\} _{i = 1}^{10} \cup \{ 2 \times i\} _{i = 1}^{10} \cup \{ 40 \times i\} _{i = 1}^{20}$.

In Algorithm 2, we set the maximum iterations $T_F=5000$ and the optimization tolerance $\epsilon_F = 10^{-5}$.
%
		\begin{figure}
			\centering
			\includegraphics[height= 220pt]{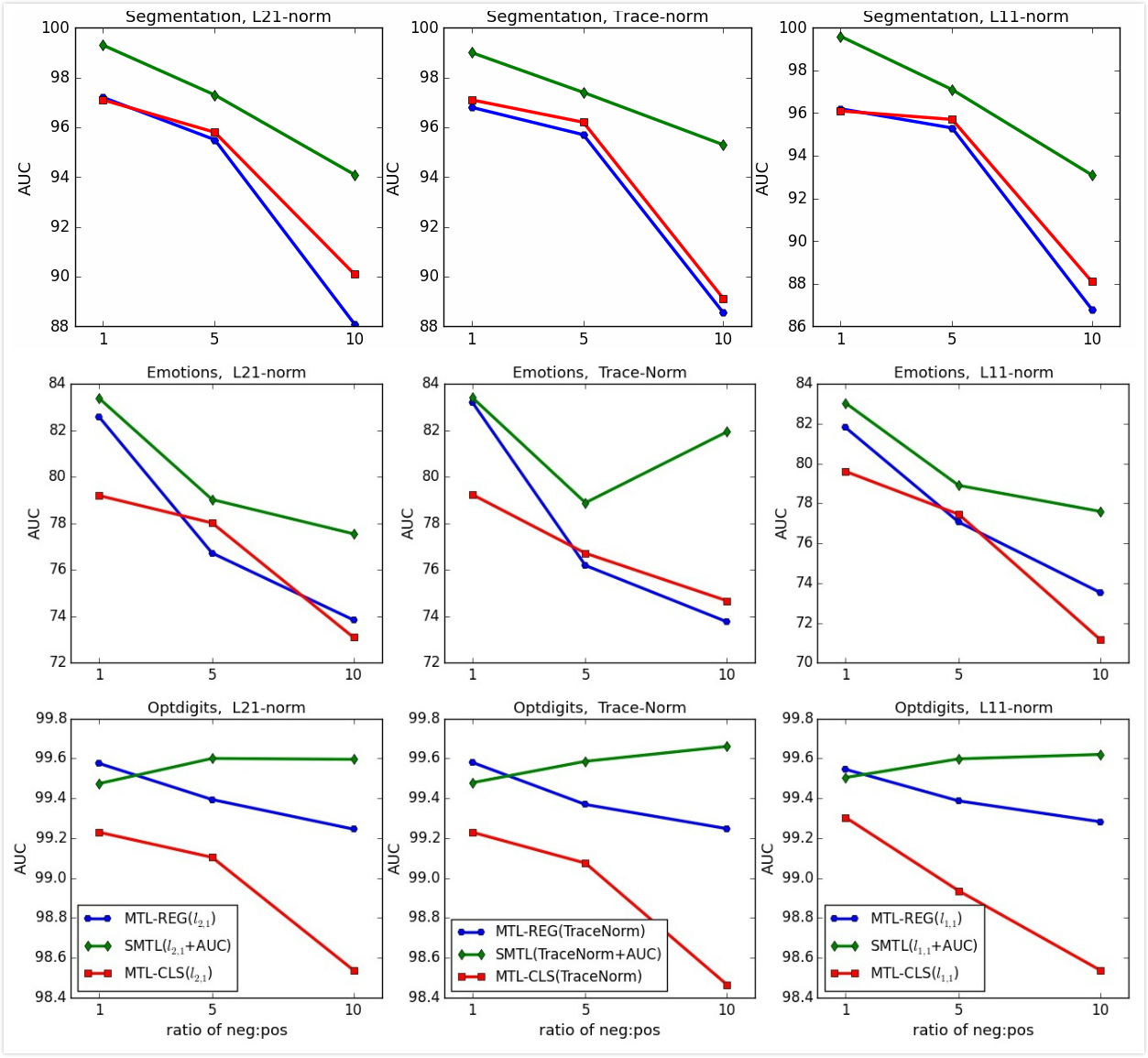}
			\caption{Comparison results on Segmentation, Emotions and Optdigits w.r.t. AUC.}
			\label{ex2_AUC}
		\end{figure}

				\begin{figure}[htbp]
					\centering
							\includegraphics[height= 220pt]{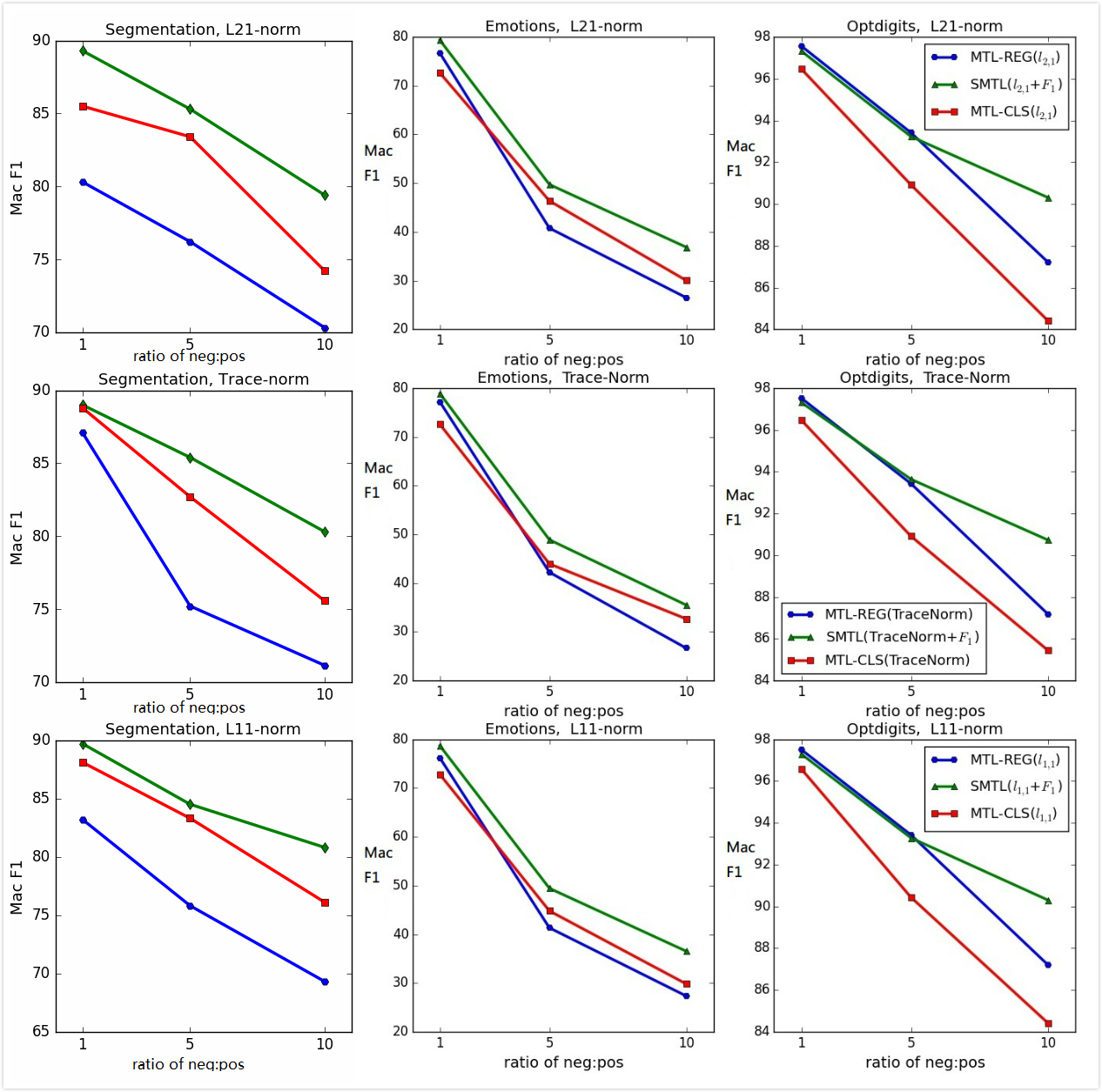}
 					\includegraphics[height= 220pt]{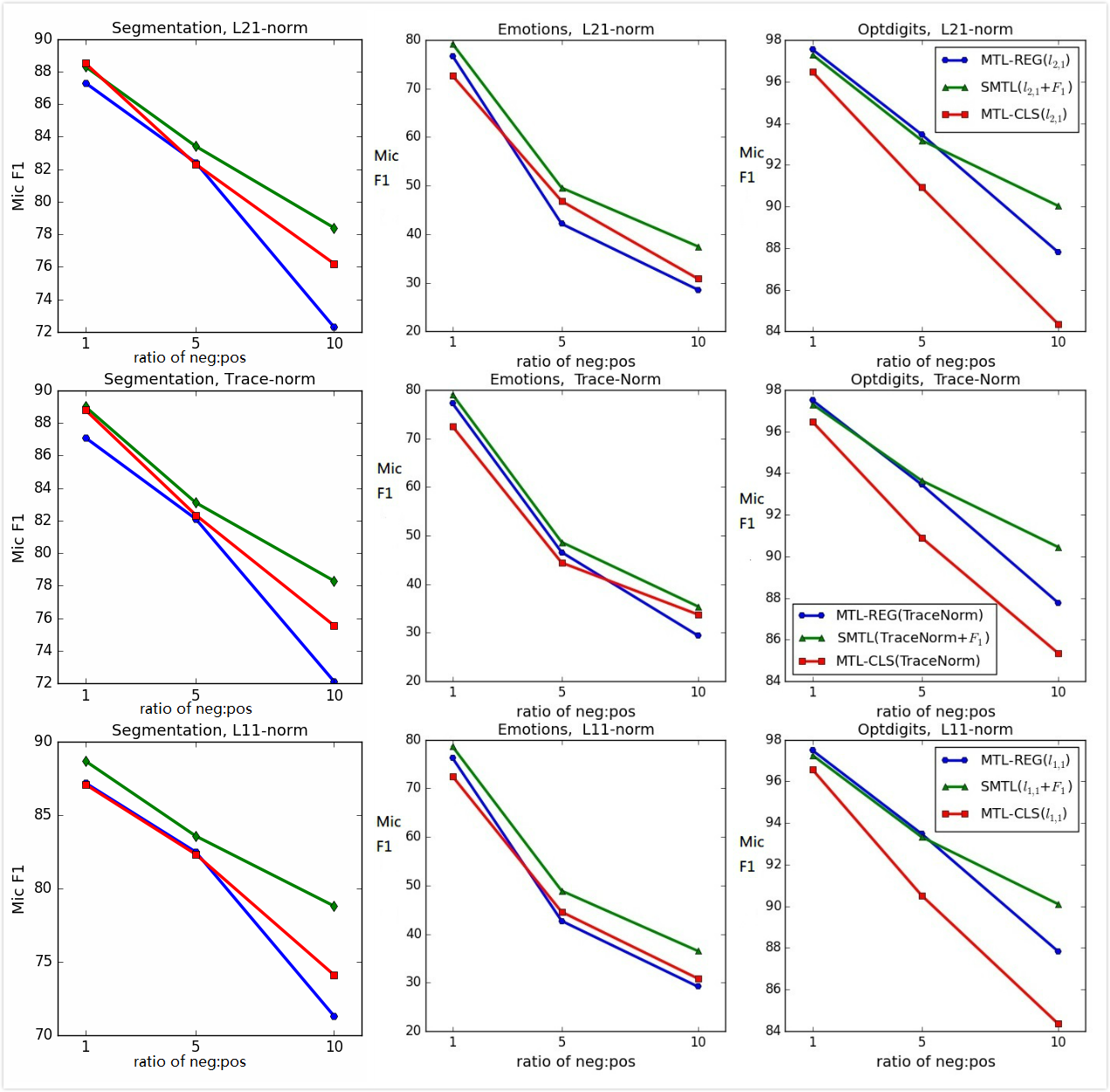}
					\caption{Comparison results on Segmentation, Emotions and Optdigits w.r.t. Macro F1 (up) and Micro F1 (down).}
					\label{ex2_F1}
				\end{figure}
\subsection{Results on real-world datasets}
The evaluation results w.r.t. Micro $F_1$, Macro $F_1$ and AUC (with standard deviations) of the proposed SMTL are shown in Table \ref{first_experiment1}, \ref{first_experiment2} and \ref{first_experiment3}. As can be seen, by using the same regularizer, the proposed SMTL variants that optimize $F_1$-score or AUC show superior performance gains over the baselines. In most cases, the SMTL variant that optimizes a specific metric achieves the best results on this metric. Here are some statistics. On the Yeast dataset,
the value of Macro $F_1$ using SMTL($\ell_{2,1}$+$F_1$) is $44.353\%$, a
$22.16\%$ relative increase compared to the best MTL baseline MTL-CLS($\ell_{2,1}$); the value of Micro $F_1$ using SMTL($\ell_{2,1}$+$F_1$) is $55.451\%$, a
$17.91\%$ relative increase compared to the best MTL baseline MTL-REG($\ell_{2,1}$); the value of averaged AUC using SMTL($\ell_{1,1}$+AUC) is $62.626\%$, a
$7.57\%$ relative increase compared to the best MTL baseline MTL-CLS($\ell_{1,1}$). On the Emotions dataset, the proposed SMTL($\ell_{2,1}$+$F_1$) performs $66.244\%$ at
Macro F1, a $4.58\%$ relative increase compared to the best MTL baseline MTL-CLS($\ell_{2,1}$); SMTL($\ell_{2,1}$+$F_1$) performs $83.378\%$ at AUC, a $2.53\%$ relative increase compared to the best MTL baseline MTL-CLS($\ell_{2,1}$); SMTL(TraceNorm+$F_1$) performs $67.6\%$ at
Macro F1, a $5.95\%$ relative increase compared to the best MTL baseline MTL-CLS(TraceNorm). On the Cal500 dataset, SMTL($\ell_{1,1}$+AUC)
performs $21.721\%$ at Macro $F_1$, compared to $12.447\%$ of MTL-REG($\ell_{1,1}$,
which indicates a $74.51\%$ relative increase; SMTL($\ell_{2,1}$+$F_1$)
performs $40.127\%$ at Micro $F_1$, compared to $37.357\%$ of MTL-CLS($\ell_{2,1}$,
which indicates a $7.41\%$ relative increase.

In addition, we conduct $t$-tests and Wilcoxon's signed rank test~\cite{wilcoxon} on $9$ datasets to investigate whether the improvements of SMTL methods against the baselines are statistically significant. The $p$-values of $t$-tests are showed in Table \ref{ttest} and \ref{ttest2}. The $p$-values of  Wilcoxon's tests are showed in Table \ref{wilcoxon_test} and \ref{wilcoxon_test2}. As can be seen, most of the $p$-values are smaller than 0.05, which indicate that the improvements are statistically significant. These results verify the effectiveness of directly optimizing evaluation metric in MTL problems.

 \begin{table}[htbp]
 	\tiny
 	\centering
 	\caption{$t$-test: $p$-values of SMTL against the baselines}
	 \begin{tabular}{|l|r|r|r|r|}
	 	\hline
	 	\multicolumn{1}{|c|}{\multirow{2}[2]{*}{Two mehtods for comparison}} &
	 	\multicolumn{1}{c|}{\multirow{2}[2]{*}{Optdigits}} &
	 	\multicolumn{1}{c|}{\multirow{2}[2]{*}{TMC2007}} &
	 	\multicolumn{1}{c|}{\multirow{2}[2]{*}{MediaMill}} &
	 	\multicolumn{1}{c|}{\multirow{2}[2]{*}{Segmentation}}
	 	\bigstrut[t]\\
	 	&
	 	&
	 	&
	 	&
	 	
	 	\bigstrut[b]\\
	 	\hline
	 	\multicolumn{5}{|c|}{Average AUC}
	 	\bigstrut\\
	 	\hline
	 	$\ell_{2,1}$: SMTL(AUC)  vs. MTL-CLS &
	 	\textbf{4.86E-07} &
	 	\textbf{1.49E-13} &
	 	\textbf{2.12E-02} &
	 	\textbf{4.74E-03}
	 	\bigstrut[t]\\
	 	$\ell_{2,1}$: SMTL(AUC)  vs. MTL-REG &
	 	\textbf{2.70E-12} &
	 	\textbf{1.44E-18} &
	 	6.58E-01 &
	 	\textbf{4.30E-03}
	 	\\
	 	Trace: SMTL(AUC)  vs. MTL-CLS &
	 	\textbf{6.88E-03} &
	 	\textbf{5.20E-03} &
	 	\textbf{2.12E-02} &
	 	\textbf{4.74E-03}
	 	\\
	 	Trace: SMTL(AUC)  vs. MTL-REG &
	 	\textbf{3.85E-12} &
	 	\textbf{4.59E-11} &
	 	6.61E-01 &
	 	\textbf{4.25E-03}
	 	\\
	 	$\ell_{1,1}$: SMTL(AUC)  vs. MTL-CLS &
	 	\textbf{5.71E-03} &
	 	\textbf{2.95E-05} &
	 	\textbf{5.52E-03} &
	 	\textbf{4.75E-03}
	 	\\
	 	$\ell_{1,1}$: SMTL(AUC)  vs. MTL-REG &
	 	\textbf{1.46E-12} &
	 	\textbf{1.92E-12} &
	 	\textbf{1.57E-08} &
	 	\textbf{4.35E-03}
	 	\\
	 	Trace: SMTL(AUC) vs. RAkEL &
	 	\textbf{1.87E-26} &
	 	\textbf{1.50E-14} &
	 	\textbf{2.79E-13} &
	 	\textbf{3.27E-14}
	 	\\
	 	Trace: SMTL(AUC) vs. MLCSSP &
	 	\textbf{6.02E-10} &
	 	\textbf{1.05E-14} &
	 	\textbf{9.63E-15} &
	 	3.24E-01
	 	\\
	 	Trace: SMTL(AUC) vs. AdaBoostMH &
	 	\textbf{2.36E-04} &
	 	\textbf{5.42E-20} &
	 	\textbf{4.44E-08} &
	 	\textbf{3.05E-14}
	 	\\
	 	Trace: SMTL(AUC) vs. HOMER &
	 	\textbf{5.23E-12} &
	 	\textbf{6.97E-09} &
	 	\textbf{4.65E-08} &
	 	\textbf{1.04E-12}
	 	\\
	 	Trace: SMTL(AUC) vs. BR &
	 	\textbf{2.91E-10} &
	 	\textbf{1.31E-16} &
	 	\textbf{2.43E-13} &
	 	\textbf{5.14E-07}
	 	\\
	 	Trace: SMTL(AUC) vs. LP &
	 	\textbf{6.82E-22} &
	 	\textbf{1.41E-20} &
	 	\textbf{1.44E-03} &
	 	9.30E-01
	 	\\
	 	Trace: SMTL(AUC) vs. ECC &
	 	\textbf{8.16E-03} &
	 	\textbf{9.67E-19} &
	 	\textbf{7.52E-03} &
	 	\textbf{6.19E-03}
	 	\bigstrut[b]\\
	 	\hline
	 	\multicolumn{5}{|c|}{Micro $F_1$}
	 	\bigstrut\\
	 	\hline
	 	$\ell_{2,1}$: SMTL($F_1$)  vs. MTL-CLS &
	 	8.37E-02 &
	 	8.37E-02 &
	 	\textbf{3.98E-10} &
	 	\textbf{4.89E-02}
	 	\bigstrut[t]\\
	 	$\ell_{2,1}$: SMTL($F_1$)  vs. MTL-REG &
	 	\textbf{3.28E-20} &
	 	\textbf{3.28E-20} &
	 	\textbf{2.54E-18} &
	 	\textbf{1.24E-02}
	 	\\
	 	Trace: SMTL($F_1$)  vs. MTL-CLS &
	 	\textbf{4.68E-03} &
	 	\textbf{4.68E-03} &
	 	\textbf{4.00E-10} &
	 	\textbf{4.96E-02}
	 	\\
	 	Trace: SMTL($F_1$)  vs. MTL-REG &
	 	\textbf{3.01E-14} &
	 	\textbf{3.01E-14} &
	 	\textbf{2.30E-18} &
	 	\textbf{4.92E-03}
	 	\\
	 	$\ell_{1,1}$: SMTL($F_1$)  vs. MTL-CLS &
	 	\textbf{9.54E-03} &
	 	\textbf{9.54E-03} &
	 	\textbf{4.19E-10} &
	 	4.75E-01
	 	\\
	 	$\ell_{1,1}$: SMTL($F_1$)  vs. MTL-REG &
	 	\textbf{6.16E-12} &
	 	\textbf{6.16E-12} &
	 	\textbf{2.56E-18} &
	 	1.03E-01
	 	\\
	 	Trace: SMTL($F_1$) vs. RAkEL &
	 	\textbf{7.30E-33} &
	 	\textbf{4.64E-25} &
	 	\textbf{4.93E-09} &
	 	\textbf{9.28E-19}
	 	\\
	 	Trace: SMTL($F_1$) vs. MLCSSP &
	 	\textbf{2.28E-30} &
	 	\textbf{1.90E-18} &
	 	\textbf{3.28E-14} &
	 	\textbf{2.90E-13}
	 	\\
	 	Trace: SMTL($F_1$) vs. AdaBoostMH &
	 	\textbf{4.53E-16} &
	 	\textbf{9.97E-35} &
	 	\textbf{9.38E-12} &
	 	\textbf{3.65E-06}
	 	\\
	 	Trace: SMTL($F_1$) vs. HOMER &
	 	\textbf{5.37E-14} &
	 	\textbf{1.13E-12} &
	 	\textbf{9.68E-12} &
	 	\textbf{8.31E-10}
	 	\\
	 	Trace: SMTL($F_1$) vs. BR &
	 	\textbf{1.61E-06} &
	 	\textbf{3.09E-14} &
	 	\textbf{5.20E-05} &
	 	\textbf{9.79E-03}
	 	\\
	 	Trace: SMTL($F_1$) vs. LP &
	 	\textbf{3.94E-20} &
	 	\textbf{9.73E-24} &
	 	\textbf{1.06E-12} &
	 	\textbf{1.39E-05}
	 	\\
	 	Trace: SMTL($F_1$) vs. ECC &
	 	\textbf{3.99E-07} &
	 	\textbf{2.76E-08} &
	 	\textbf{1.75E-16} &
	 	1.45E-01
	 	\bigstrut[b]\\
	 	\hline
	 	\multicolumn{5}{|c|}{Macro $F_1$}
	 	\bigstrut\\
	 	\hline
	 	$\ell_{2,1}$: SMTL($F_1$)  vs. MTL-CLS &
	 	\textbf{4.09E-21} &
	 	\textbf{1.61E-10} &
	 	\textbf{3.98E-10} &
	 	\textbf{4.09E-02}
	 	\bigstrut[t]\\
	 	$\ell_{2,1}$: SMTL($F_1$)  vs. MTL-REG &
	 	\textbf{1.47E-26} &
	 	\textbf{3.09E-16} &
	 	\textbf{2.54E-18} &
	 	\textbf{2.98E-12}
	 	\\
	 	Trace: SMTL($F_1$)  vs. MTL-CLS &
	 	\textbf{1.04E-21} &
	 	1.82E-02 &
	 	\textbf{4.00E-10} &
	 	\textbf{4.13E-02}
	 	\\
	 	Trace: SMTL($F_1$)  vs. MTL-REG &
	 	\textbf{3.85E-19} &
	 	\textbf{5.87E-12} &
	 	\textbf{2.30E-18} &
	 	\textbf{3.19E-12}
	 	\\
	 	$\ell_{1,1}$: SMTL($F_1$)  vs. MTL-CLS &
	 	\textbf{7.04E-22} &
	 	\textbf{9.26E-07} &
	 	\textbf{4.19E-10} &
	 	\textbf{4.04E-02}
	 	\\
	 	$\ell_{1,1}$: SMTL($F_1$)  vs. MTL-REG &
	 	\textbf{1.94E-24} &
	 	\textbf{8.74E-14} &
	 	\textbf{2.56E-18} &
	 	\textbf{2.57E-12}
	 	\\
	 	Trace: SMTL($F_1$) vs. RAkEL &
	 	\textbf{6.35E-29} &
	 	\textbf{3.99E-10} &
	 	\textbf{4.93E-09} &
	 	\textbf{3.08E-16}
	 	\\
	 	Trace: SMTL($F_1$) vs. MLCSSP &
	 	\textbf{6.50E-16} &
	 	\textbf{2.29E-10} &
	 	\textbf{3.28E-14} &
	 	\textbf{4.68E-02}
	 	\\
	 	Trace: SMTL($F_1$) vs. AdaBoostMH &
	 	\textbf{1.21E-04} &
	 	\textbf{2.99E-23} &
	 	\textbf{9.38E-12} &
	 	\textbf{3.17E-16}
	 	\\
	 	Trace: SMTL($F_1$) vs. HOMER &
	 	\textbf{1.78E-11} &
	 	\textbf{4.33E-02} &
	 	\textbf{9.68E-12} &
	 	\textbf{2.55E-11}
	 	\\
	 	Trace: SMTL($F_1$) vs. BR &
	 	\textbf{1.28E-08} &
	 	\textbf{1.12E-13} &
	 	\textbf{5.20E-05} &
	 	\textbf{1.19E-02}
	 	\\
	 	Trace: SMTL($F_1$) vs. LP &
	 	\textbf{1.67E-19} &
	 	\textbf{1.52E-14} &
	 	\textbf{1.06E-12} &
	 	7.49E-01
	 	\\
	 	Trace: SMTL($F_1$) vs. ECC &
	 	2.83E-01 &
	 	\textbf{4.46E-08} &
	 	\textbf{1.75E-16} &
	 	9.52E-01
	 	\bigstrut[b]\\
	 	\hline
	 \end{tabular}%
 	\label{ttest}%
 \end{table}%

 \begin{table}[htbp]
 	\tiny
 	\centering
 	\caption{$t$-test: $p$-values of SMTL against the baselines}
	\begin{tabular}{|l|r|r|r|r|r|}
		\hline
		\multicolumn{1}{|c|}{\multirow{2}[2]{*}{Two mehtods for comparison}} &
		\multicolumn{1}{c|}{\multirow{2}[2]{*}{Cal500}} &
		\multicolumn{1}{c|}{\multirow{2}[2]{*}{Yeast}} &
		\multicolumn{1}{c|}{\multirow{2}[2]{*}{Emotions}} &
		\multicolumn{1}{c|}{\multirow{2}[2]{*}{Scene}} &
		\multicolumn{1}{c|}{\multirow{2}[2]{*}{Flags}}
		\bigstrut[t]\\
		&
		&
		&
		&
		&
		
		\bigstrut[b]\\
		\hline
		\multicolumn{6}{|c|}{Average AUC}
		\bigstrut\\
		\hline
		$\ell_{2,1}$: SMTL(AUC)  vs. MTL-CLS &
		\textbf{2.62E-01} &
		\textbf{1.02E-12} &
		2.62E-01 &
		\textbf{4.92E-02} &
		\textbf{4.30E-02}
		\bigstrut[t]\\
		$\ell_{2,1}$: SMTL(AUC)  vs. MTL-REG &
		\textbf{7.48E-05} &
		\textbf{1.47E-09} &
		\textbf{7.48E-05} &
		\textbf{4.74E-11} &
		\textbf{4.21E-02}
		\\
		Trace: SMTL(AUC)  vs. MTL-CLS &
		\textbf{1.01E-01} &
		\textbf{8.97E-13} &
		1.01E-01 &
		\textbf{4.48E-02} &
		\textbf{4.21E-02}
		\\
		Trace: SMTL(AUC)  vs. MTL-REG &
		\textbf{3.04E-03} &
		\textbf{1.53E-09} &
		\textbf{3.04E-03} &
		\textbf{5.00E-11} &
		\textbf{4.37E-02}
		\\
		$\ell_{1,1}$: SMTL(AUC)  vs. MTL-CLS &
		\textbf{2.18E-03} &
		\textbf{1.00E-12} &
		\textbf{2.18E-03} &
		\textbf{4.56E-02} &
		\textbf{4.27E-02}
		\\
		$\ell_{1,1}$: SMTL(AUC)  vs. MTL-REG &
		\textbf{2.55E-06} &
		\textbf{1.71E-09} &
		\textbf{2.55E-06} &
		\textbf{4.67E-11} &
		\textbf{4.28E-02}
		\\
		Trace: SMTL(AUC) vs. RAkEL &
		\textbf{2.62E-12} &
		\textbf{1.65E-10} &
		\textbf{4.55E-04} &
		1.48E-01 &
		\textbf{5.58E-05}
		\\
		Trace: SMTL(AUC) vs. MLCSSP &
		\textbf{1.49E-21} &
		\textbf{1.05E-07} &
		\textbf{1.22E-04} &
		\textbf{1.60E-15} &
		\textbf{6.81E-11}
		\\
		Trace: SMTL(AUC) vs. AdaBoostMH &
		\textbf{4.10E-33} &
		\textbf{1.03E-06} &
		\textbf{3.61E-23} &
		\textbf{3.21E-19} &
		\textbf{2.12E-02}
		\\
		Trace: SMTL(AUC) vs. HOMER &
		\textbf{2.30E-13} &
		\textbf{2.54E-04} &
		\textbf{8.57E-09} &
		\textbf{4.33E-02} &
		\textbf{9.89E-09}
		\\
		Trace: SMTL(AUC) vs. BR &
		\textbf{2.23E-16} &
		\textbf{7.57E-10} &
		\textbf{3.84E-06} &
		7.92E-02 &
		\textbf{1.63E-06}
		\\
		Trace: SMTL(AUC) vs. LP &
		\textbf{1.12E-14} &
		\textbf{6.42E-07} &
		\textbf{9.15E-15} &
		4.97E-01 &
		\textbf{3.19E-03}
		\\
		Trace: SMTL(AUC) vs. ECC &
		\textbf{2.00E-13} &
		\textbf{2.09E-12} &
		\textbf{6.45E-07} &
		3.28E-01 &
		9.78E-01
		\bigstrut[b]\\
		\hline
		\multicolumn{6}{|c|}{Micro $F_1$}
		\bigstrut\\
		\hline
		$\ell_{2,1}$: SMTL($F_1$)  vs. MTL-CLS &
		\textbf{4.09E-21} &
		\textbf{2.70E-05} &
		2.62E-01 &
		\textbf{1.64E-05} &
		\textbf{3.14E-02}
		\bigstrut[t]\\
		$\ell_{2,1}$: SMTL($F_1$)  vs. MTL-REG &
		\textbf{1.47E-26} &
		\textbf{5.00E-02} &
		\textbf{7.48E-05} &
		\textbf{1.09E-06} &
		\textbf{1.87E-03}
		\\
		Trace: SMTL($F_1$)  vs. MTL-CLS &
		\textbf{1.04E-21} &
		\textbf{3.45E-05} &
		1.01E-01 &
		\textbf{1.42E-05} &
		\textbf{3.10E-02}
		\\
		Trace: SMTL($F_1$)  vs. MTL-REG &
		\textbf{3.85E-19} &
		\textbf{4.39E-02} &
		\textbf{3.04E-03} &
		\textbf{1.19E-06} &
		\textbf{1.76E-03}
		\\
		$\ell_{1,1}$: SMTL($F_1$)  vs. MTL-CLS &
		\textbf{7.04E-22} &
		\textbf{2.16E-05} &
		\textbf{2.18E-03} &
		\textbf{1.54E-05} &
		\textbf{3.13E-02}
		\\
		$\ell_{1,1}$: SMTL($F_1$)  vs. MTL-REG &
		\textbf{1.94E-24} &
		\textbf{4.21E-02} &
		\textbf{2.55E-06} &
		\textbf{1.35E-06} &
		\textbf{1.87E-03}
		\\
		Trace: SMTL($F_1$) vs. RAkEL &
		\textbf{4.16E-08} &
		\textbf{2.54E-03} &
		\textbf{4.55E-04} &
		\textbf{3.26E-15} &
		\textbf{2.85E-08}
		\\
		Trace: SMTL($F_1$) vs. MLCSSP &
		\textbf{2.82E-10} &
		\textbf{9.31E-21} &
		\textbf{1.22E-04} &
		\textbf{1.80E-16} &
		\textbf{2.01E-13}
		\\
		Trace: SMTL($F_1$) vs. AdaBoostMH &
		\textbf{8.68E-17} &
		\textbf{3.36E-22} &
		\textbf{3.61E-23} &
		\textbf{4.76E-02} &
		\textbf{7.77E-05}
		\\
		Trace: SMTL($F_1$) vs. HOMER &
		\textbf{5.69E-11} &
		1.08E-01 &
		\textbf{8.57E-09} &
		\textbf{4.53E-14} &
		\textbf{3.25E-10}
		\\
		Trace: SMTL($F_1$) vs. BR &
		\textbf{7.35E-21} &
		1.08E-01 &
		\textbf{3.84E-06} &
		\textbf{2.51E-09} &
		\textbf{4.96E-06}
		\\
		Trace: SMTL($F_1$) vs. LP &
		\textbf{1.64E-13} &
		\textbf{9.74E-11} &
		\textbf{9.15E-15} &
		\textbf{1.25E-14} &
		\textbf{3.39E-08}
		\\
		Trace: SMTL($F_1$) vs. ECC &
		\textbf{6.21E-14} &
		5.30E-01 &
		\textbf{6.45E-07} &
		\textbf{1.05E-02} &
		9.09E-01
		\bigstrut[b]\\
		\hline
		\multicolumn{6}{|c|}{Macro $F_1$}
		\bigstrut\\
		\hline
		$\ell_{2,1}$: SMTL($F_1$)  vs. MTL-CLS &
		2.43E-02 &
		\textbf{2.51E-06} &
		\textbf{7.55E-12} &
		\textbf{4.09E-02} &
		\textbf{1.12E-02}
		\bigstrut[t]\\
		$\ell_{2,1}$: SMTL($F_1$)  vs. MTL-REG &
		\textbf{4.24E-10} &
		\textbf{2.71E-19} &
		\textbf{2.83E-09} &
		\textbf{1.40E-10} &
		\textbf{2.58E-03}
		\\
		Trace: SMTL($F_1$)  vs. MTL-CLS &
		\textbf{1.77E-05} &
		\textbf{2.34E-06} &
		\textbf{3.53E-09} &
		\textbf{4.37E-02} &
		\textbf{1.13E-02}
		\\
		Trace: SMTL($F_1$)  vs. MTL-REG &
		\textbf{1.43E-04} &
		\textbf{3.33E-19} &
		\textbf{2.69E-02} &
		\textbf{1.55E-10} &
		\textbf{2.67E-03}
		\\
		$\ell_{1,1}$: SMTL($F_1$)  vs. MTL-CLS &
		3.99E-02 &
		\textbf{2.66E-06} &
		\textbf{6.38E-12} &
		\textbf{4.43E-02} &
		\textbf{1.11E-02}
		\\
		$\ell_{1,1}$: SMTL($F_1$)  vs. MTL-REG &
		\textbf{1.01E-12} &
		\textbf{2.76E-19} &
		\textbf{1.09E-06} &
		\textbf{1.32E-10} &
		\textbf{2.54E-03}
		\\
		Trace: SMTL($F_1$) vs. RAkEL &
		\textbf{5.45E-05} &
		\textbf{5.52E-03} &
		\textbf{3.24E-05} &
		5.68E-02 &
		\textbf{2.11E-04}
		\\
		Trace: SMTL($F_1$) vs. MLCSSP &
		6.64E-01 &
		\textbf{1.57E-08} &
		\textbf{6.89E-05} &
		\textbf{2.34E-18} &
		\textbf{2.75E-10}
		\\
		Trace: SMTL($F_1$) vs. AdaBoostMH &
		\textbf{1.28E-29} &
		\textbf{1.36E-28} &
		\textbf{3.03E-28} &
		\textbf{5.89E-21} &
		\textbf{1.16E-07}
		\\
		Trace: SMTL($F_1$) vs. HOMER &
		\textbf{3.52E-23} &
		\textbf{6.16E-10} &
		\textbf{2.60E-09} &
		\textbf{3.77E-06} &
		\textbf{1.84E-11}
		\\
		Trace: SMTL($F_1$) vs. BR &
		\textbf{7.28E-14} &
		\textbf{6.97E-12} &
		\textbf{6.40E-07} &
		\textbf{2.49E-01} &
		\textbf{2.86E-09}
		\\
		Trace: SMTL($F_1$) vs. LP &
		\textbf{1.42E-18} &
		\textbf{9.45E-16} &
		\textbf{7.63E-15} &
		4.04E-01 &
		\textbf{5.68E-05}
		\\
		Trace: SMTL($F_1$) vs. ECC &
		\textbf{5.69E-21} &
		\textbf{1.98E-16} &
		\textbf{5.51E-08} &
		9.15E-01 &
		\textbf{4.96E-03}
		\bigstrut[b]\\
		\hline
	\end{tabular}%
	
 	\label{ttest2}%
 \end{table}%

 \begin{table}[htbp]
 	\tiny
 	\centering
 	\caption{Wilcoxon's test: $p$-values of SMTL against the baselines}
 	\begin{tabular}{|l|r|r|r|r|}
 		\hline
 		\multicolumn{1}{|c|}{\multirow{2}[2]{*}{Two mehtods for comparison}} &
 		\multicolumn{1}{c|}{\multirow{2}[2]{*}{Optdigits}} &
 		\multicolumn{1}{c|}{\multirow{2}[2]{*}{TMC2007}} &
 		\multicolumn{1}{c|}{\multirow{2}[2]{*}{MediaMill}} &
 		\multicolumn{1}{c|}{\multirow{2}[2]{*}{Segmentation}}
 		\bigstrut[t]\\
 		&
 		&
 		&
 		&
 		
 		\bigstrut[b]\\
 		\hline
 		\multicolumn{5}{|c|}{Average AUC}
 		\bigstrut\\
 		\hline
 		$\ell_{2,1}$: SMTL(AUC)  vs. MTL-CLS &
 		\textbf{1.25E-02} &
 		\textbf{1.25E-02} &
 		\textbf{5.06E-03} &
 		5.75E-01
 		\\
 		$\ell_{2,1}$: SMTL(AUC)  vs. MTL-REG &
 		\textbf{5.06E-03} &
 		\textbf{5.06E-03} &
 		4.45E-01 &
 		\textbf{2.84E-02}
 		\\
 		Trace: SMTL(AUC)  vs. MTL-CLS &
 		\textbf{2.84E-02} &
 		\textbf{2.18E-02} &
 		\textbf{2.18E-02} &
 		\textbf{4.69E-02}
 		\\
 		Trace: SMTL(AUC)  vs. MTL-REG &
 		\textbf{5.06E-03} &
 		\textbf{5.06E-03} &
 		8.79E-01 &
 		3.86E-01
 		\\
 		$\ell_{1,1}$: SMTL(AUC)  vs. MTL-CLS &
 		\textbf{2.84E-02} &
 		\textbf{2.84E-02} &
 		\textbf{4.69E-02} &
 		\textbf{2.18E-02}
 		\\
 		$\ell_{1,1}$: SMTL(AUC)  vs. MTL-REG &
 		\textbf{5.06E-03} &
 		\textbf{5.06E-03} &
 		5.75E-01 &
 		\textbf{2.18E-02}
 		\\
 		Trace: SMTL(AUC) vs. RAkEL &
 		\textbf{5.06E-03} &
 		\textbf{5.06E-03} &
 		\textbf{5.06E-03} &
 		\textbf{5.06E-03}
 		\\
 		Trace: SMTL(AUC) vs. MLCSSP &
 		\textbf{5.06E-03} &
 		\textbf{5.06E-03} &
 		\textbf{5.06E-03} &
 		2.85E-01
 		\\
 		Trace: SMTL(AUC) vs. AdaBoostMH &
 		\textbf{5.06E-03} &
 		\textbf{5.06E-03} &
 		\textbf{5.06E-03} &
 		\textbf{5.06E-03}
 		\\
 		Trace: SMTL(AUC) vs. HOMER &
 		\textbf{5.06E-03} &
 		\textbf{5.06E-03} &
 		\textbf{5.06E-03} &
 		\textbf{5.06E-03}
 		\\
 		Trace: SMTL(AUC) vs. BR &
 		\textbf{5.06E-03} &
 		\textbf{5.06E-03} &
 		\textbf{5.06E-03} &
 		\textbf{5.06E-03}
 		\\
 		Trace: SMTL(AUC) vs. LP &
 		\textbf{5.06E-03} &
 		\textbf{5.06E-03} &
 		\textbf{1.25E-02} &
 		8.79E-01
 		\\
 		Trace: SMTL(AUC) vs. ECC &
 		7.45E-02 &
 		\textbf{5.06E-03} &
 		\textbf{3.67E-02} &
 		\textbf{1.66E-02}
 		\bigstrut[b]\\
 		\hline
 		\multicolumn{5}{|c|}{Micro $F_1$}
 		\bigstrut\\
 		\hline
 		$\ell_{2,1}$: SMTL($F_1$)  vs. MTL-CLS &
 		\textbf{5.06E-03} &
 		5.93E-02 &
 		\textbf{5.06E-03} &
 		\textbf{9.34E-03}
 		\bigstrut[t]\\
 		$\ell_{2,1}$: SMTL($F_1$)  vs. MTL-REG &
 		\textbf{5.06E-03} &
 		\textbf{5.06E-03} &
 		\textbf{5.06E-03} &
 		\textbf{2.84E-02}
 		\\
 		Trace: SMTL($F_1$)  vs. MTL-CLS &
 		\textbf{5.06E-03} &
 		\textbf{1.66E-02} &
 		\textbf{5.06E-03} &
 		9.26E-02
 		\\
 		Trace: SMTL($F_1$)  vs. MTL-REG &
 		\textbf{5.06E-03} &
 		\textbf{5.06E-03} &
 		\textbf{5.06E-03} &
 		\textbf{6.91E-03}
 		\\
 		$\ell_{1,1}$: SMTL($F_1$)  vs. MTL-CLS &
 		\textbf{5.06E-03} &
 		\textbf{4.69E-02} &
 		\textbf{5.06E-03} &
 		5.93E-02
 		\\
 		$\ell_{1,1}$: SMTL($F_1$)  vs. MTL-REG &
 		\textbf{5.06E-03} &
 		\textbf{5.06E-03} &
 		\textbf{5.06E-03} &
 		\textbf{2.84E-02}
 		\\
 		Trace: SMTL($F_1$) vs. RAkEL &
 		\textbf{5.06E-03} &
 		\textbf{5.06E-03} &
 		\textbf{5.06E-03} &
 		\textbf{5.06E-03}
 		\\
 		Trace: SMTL($F_1$) vs. MLCSSP &
 		\textbf{5.06E-03} &
 		\textbf{5.06E-03} &
 		\textbf{5.06E-03} &
 		\textbf{5.06E-03}
 		\\
 		Trace: SMTL($F_1$) vs. AdaBoostMH &
 		\textbf{5.06E-03} &
 		\textbf{5.06E-03} &
 		\textbf{5.06E-03} &
 		\textbf{5.06E-03}
 		\\
 		Trace: SMTL($F_1$) vs. HOMER &
 		\textbf{5.06E-03} &
 		\textbf{5.06E-03} &
 		\textbf{5.06E-03} &
 		\textbf{5.06E-03}
 		\\
 		Trace: SMTL(($F_1$) vs. BR &
 		\textbf{5.06E-03} &
 		\textbf{5.06E-03} &
 		\textbf{9.34E-03} &
 		1.69E-01
 		\\
 		Trace: SMTL(($F_1$) vs. LP &
 		\textbf{5.06E-03} &
 		\textbf{5.06E-03} &
 		\textbf{5.06E-03} &
 		\textbf{5.06E-03}
 		\\
 		Trace: SMTL(($F_1$) vs. ECC &
 		\textbf{5.06E-03} &
 		\textbf{5.06E-03} &
 		\textbf{5.06E-03} &
 		2.03E-01
 		\bigstrut[b]\\
 		\hline
 		\multicolumn{5}{|c|}{Macro $F_1$}
 		\bigstrut\\
 		\hline
 		$\ell_{2,1}$: SMTL($F_1$)  vs. MTL-CLS &
 		\textbf{9.34E-03} &
 		\textbf{5.06E-03} &
 		\textbf{5.06E-03} &
 		\textbf{4.69E-02}
 		\bigstrut[t]\\
 		$\ell_{2,1}$: SMTL($F_1$)  vs. MTL-REG &
 		\textbf{5.06E-03} &
 		\textbf{5.06E-03} &
 		\textbf{5.06E-03} &
 		\textbf{5.06E-03}
 		\\
 		Trace: SMTL($F_1$)  vs. MTL-CLS &
 		\textbf{9.34E-03} &
 		\textbf{5.06E-03} &
 		\textbf{5.06E-03} &
 		5.93E-02
 		\\
 		Trace: SMTL($F_1$)  vs. MTL-REG &
 		\textbf{5.06E-03} &
 		\textbf{5.06E-03} &
 		\textbf{5.06E-03} &
 		\textbf{5.06E-03}
 		\\
 		$\ell_{1,1}$: SMTL($F_1$)  vs. MTL-CLS &
 		\textbf{9.34E-03} &
 		\textbf{5.06E-03} &
 		\textbf{5.06E-03} &
 		1.14E-01
 		\\
 		$\ell_{1,1}$: SMTL($F_1$)  vs. MTL-REG &
 		\textbf{5.06E-03} &
 		\textbf{5.06E-03} &
 		\textbf{5.06E-03} &
 		\textbf{5.06E-03}
 		\\
 		Trace: SMTL($F_1$) vs. RAkEL &
 		\textbf{5.06E-03} &
 		\textbf{5.06E-03} &
 		\textbf{5.06E-03} &
 		\textbf{5.06E-03}
 		\\
 		Trace: SMTL($F_1$) vs. MLCSSP &
 		\textbf{5.06E-03} &
 		\textbf{5.06E-03} &
 		\textbf{5.06E-03} &
 		7.45E-02
 		\\
 		Trace: SMTL($F_1$) vs. AdaBoostMH &
 		\textbf{5.06E-03} &
 		\textbf{5.06E-03} &
 		\textbf{5.06E-03} &
 		\textbf{5.06E-03}
 		\\
 		Trace: SMTL($F_1$) vs. HOMER &
 		\textbf{5.06E-03} &
 		\textbf{3.67E-02} &
 		\textbf{5.06E-03} &
 		\textbf{5.06E-03}
 		\\
 		Trace: SMTL(($F_1$) vs. BR &
 		\textbf{5.06E-03} &
 		\textbf{5.06E-03} &
 		\textbf{5.06E-03} &
 		\textbf{2.84E-02}
 		\\
 		Trace: SMTL(($F_1$) vs. LP &
 		\textbf{5.06E-03} &
 		\textbf{5.06E-03} &
 		1.69E-01 &
 		8.79E-01
 		\\
 		Trace: SMTL(($F_1$) vs. ECC &
 		\textbf{4.69E-02} &
 		\textbf{5.06E-03} &
 		\textbf{1.66E-02} &
 		9.59E-01
 		\bigstrut[b]\\
 		\hline
 	\end{tabular}%
 	
 	\label{wilcoxon_test}%
 \end{table}%

 \begin{table}[htbp]
 	\tiny
 	\centering
 	\caption{Wilcoxon's test: $p$-values of SMTL against the baselines}
 	\begin{tabular}{|l|r|r|r|r|r|}
 		\hline
 		\multicolumn{1}{|c|}{\multirow{2}[2]{*}{Two mehtods for comparison}} &
 		\multicolumn{1}{c|}{\multirow{2}[2]{*}{Cal500}} &
 		\multicolumn{1}{c|}{\multirow{2}[2]{*}{Yeast}} &
 		\multicolumn{1}{c|}{\multirow{2}[2]{*}{Emotions}} &
 		\multicolumn{1}{c|}{\multirow{2}[2]{*}{Scene}} &
 		\multicolumn{1}{c|}{\multirow{2}[2]{*}{Flags}}
 		\bigstrut[t]\\
 		&
 		&
 		&
 		&
 		&
 		
 		\bigstrut[b]\\
 		\hline
 		\multicolumn{6}{|c|}{Average AUC}
 		\bigstrut\\
 		\hline
 		$\ell_{2,1}$: SMTL(AUC)  vs. MTL-CLS &
 		\textbf{5.06E-03} &
 		\textbf{5.06E-03} &
 		1.69E-01 &
 		\textbf{1.25E-02} &
 		\textbf{1.25E-02}
 		\bigstrut[t]\\
 		$\ell_{2,1}$: SMTL(AUC)  vs. MTL-REG &
 		\textbf{5.06E-03} &
 		\textbf{5.06E-03} &
 		\textbf{1.25E-02} &
 		\textbf{5.06E-03} &
 		\textbf{5.06E-03}
 		\\
 		Trace: SMTL(AUC)  vs. MTL-CLS &
 		\textbf{6.91E-03} &
 		\textbf{5.06E-03} &
 		1.14E-01 &
 		\textbf{4.69E-02} &
 		5.08E-01
 		\\
 		Trace: SMTL(AUC)  vs. MTL-REG &
 		\textbf{5.06E-03} &
 		\textbf{5.06E-03} &
 		\textbf{1.25E-02} &
 		\textbf{5.06E-03} &
 		\textbf{3.67E-02}
 		\\
 		$\ell_{1,1}$: SMTL(AUC)  vs. MTL-CLS &
 		\textbf{5.06E-03} &
 		\textbf{5.06E-03} &
 		\textbf{1.25E-02} &
 		\textbf{2.84E-02} &
 		\textbf{1.25E-02}
 		\\
 		$\ell_{1,1}$: SMTL(AUC)  vs. MTL-REG &
 		\textbf{5.06E-03} &
 		\textbf{5.06E-03} &
 		\textbf{1.25E-02} &
 		\textbf{5.06E-03} &
 		\textbf{1.25E-02}
 		\\
 		Trace: SMTL(AUC) vs. RAkEL &
 		\textbf{5.06E-03} &
 		\textbf{5.06E-03} &
 		\textbf{1.25E-02} &
 		\textbf{2.84E-02} &
 		\textbf{5.06E-03}
 		\\
 		Trace: SMTL(AUC) vs. MLCSSP &
 		\textbf{5.06E-03} &
 		\textbf{5.06E-03} &
 		\textbf{5.06E-03} &
 		\textbf{5.06E-03} &
 		\textbf{5.06E-03}
 		\\
 		Trace: SMTL(AUC) vs. AdaBoostMH &
 		\textbf{5.06E-03} &
 		\textbf{5.06E-03} &
 		\textbf{5.06E-03} &
 		\textbf{5.06E-03} &
 		\textbf{1.25E-02}
 		\\
 		Trace: SMTL(AUC) vs. HOMER &
 		\textbf{5.06E-03} &
 		\textbf{5.06E-03} &
 		\textbf{5.06E-03} &
 		\textbf{5.06E-03} &
 		\textbf{5.06E-03}
 		\\
 		Trace: SMTL(AUC) vs. BR &
 		\textbf{5.06E-03} &
 		\textbf{5.06E-03} &
 		\textbf{5.06E-03} &
 		9.26E-02 &
 		\textbf{5.06E-03}
 		\\
 		Trace: SMTL(AUC) vs. LP &
 		\textbf{5.06E-03} &
 		\textbf{6.91E-03} &
 		\textbf{5.06E-03} &
 		7.21E-01 &
 		\textbf{9.34E-03}
 		\\
 		Trace: SMTL(AUC) vs. ECC &
 		\textbf{5.06E-03} &
 		\textbf{5.06E-03} &
 		\textbf{5.06E-03} &
 		1.69E-01 &
 		9.59E-01
 		\bigstrut[b]\\
 		\hline
 		\multicolumn{6}{|c|}{Micro $F_1$}
 		\bigstrut\\
 		\hline
 		$\ell_{2,1}$: SMTL($F_1$)  vs. MTL-CLS &
 		\textbf{5.06E-03} &
 		\textbf{6.91E-03} &
 		\textbf{5.06E-03} &
 		\textbf{5.06E-03} &
 		\textbf{5.06E-03}
 		\bigstrut[t]\\
 		$\ell_{2,1}$: SMTL($F_1$)  vs. MTL-REG &
 		\textbf{9.34E-03} &
 		\textbf{2.84E-02} &
 		\textbf{2.84E-02} &
 		\textbf{5.06E-03} &
 		\textbf{1.25E-02}
 		\\
 		Trace: SMTL($F_1$)  vs. MTL-CLS &
 		\textbf{5.06E-03} &
 		\textbf{6.91E-03} &
 		\textbf{5.06E-03} &
 		\textbf{5.06E-03} &
 		4.45E-01
 		\\
 		Trace: SMTL($F_1$)  vs. MTL-REG &
 		\textbf{5.06E-03} &
 		\textbf{2.84E-02} &
 		\textbf{1.25E-02} &
 		\textbf{5.06E-03} &
 		\textbf{1.25E-02}
 		\\
 		$\ell_{1,1}$: SMTL($F_1$)  vs. MTL-CLS &
 		\textbf{5.06E-03} &
 		\textbf{5.06E-03} &
 		\textbf{5.06E-03} &
 		\textbf{5.06E-03} &
 		\textbf{3.67E-02}
 		\\
 		$\ell_{1,1}$: SMTL($F_1$)  vs. MTL-REG &
 		\textbf{5.06E-03} &
 		\textbf{2.84E-02} &
 		\textbf{3.67E-02} &
 		\textbf{5.06E-03} &
 		\textbf{2.84E-02}
 		\\
 		Trace: SMTL($F_1$) vs. RAkEL &
 		\textbf{5.06E-03} &
 		\textbf{9.34E-03} &
 		\textbf{5.06E-03} &
 		\textbf{5.06E-03} &
 		\textbf{5.06E-03}
 		\\
 		Trace: SMTL($F_1$) vs. MLCSSP &
 		\textbf{5.06E-03} &
 		\textbf{5.06E-03} &
 		\textbf{5.06E-03} &
 		\textbf{5.06E-03} &
 		\textbf{5.06E-03}
 		\\
 		Trace: SMTL($F_1$) vs. AdaBoostMH &
 		\textbf{5.06E-03} &
 		\textbf{5.06E-03} &
 		\textbf{5.06E-03} &
 		5.93E-02 &
 		\textbf{5.06E-03}
 		\\
 		Trace: SMTL($F_1$) vs. HOMER &
 		\textbf{5.06E-03} &
 		2.03E-01 &
 		\textbf{5.06E-03} &
 		\textbf{5.06E-03} &
 		\textbf{5.06E-03}
 		\\
 		Trace: SMTL(($F_1$) vs. BR &
 		\textbf{5.06E-03} &
 		\textbf{5.06E-03} &
 		\textbf{5.06E-03} &
 		\textbf{5.06E-03} &
 		\textbf{5.06E-03}
 		\\
 		Trace: SMTL(($F_1$) vs. LP &
 		\textbf{5.06E-03} &
 		\textbf{5.06E-03} &
 		\textbf{5.06E-03} &
 		\textbf{5.06E-03} &
 		\textbf{5.06E-03}
 		\\
 		Trace: SMTL(($F_1$) vs. ECC &
 		\textbf{5.06E-03} &
 		3.86E-01 &
 		2.41E-01 &
 		\textbf{4.69E-02} &
 		8.79E-01
 		\bigstrut[b]\\
 		\hline
 		\multicolumn{6}{|c|}{Macro $F_1$}
 		\bigstrut\\
 		\hline
 		$\ell_{2,1}$: SMTL($F_1$)  vs. MTL-CLS &
 		\textbf{5.06E-03} &
 		\textbf{5.06E-03} &
 		\textbf{6.91E-03} &
 		\textbf{3.67E-02} &
 		\textbf{3.67E-02}
 		\bigstrut[t]\\
 		$\ell_{2,1}$: SMTL($F_1$)  vs. MTL-REG &
 		\textbf{5.06E-03} &
 		\textbf{5.06E-03} &
 		\textbf{5.06E-03} &
 		\textbf{5.06E-03} &
 		\textbf{6.91E-03}
 		\\
 		Trace: SMTL($F_1$)  vs. MTL-CLS &
 		\textbf{5.06E-03} &
 		\textbf{5.06E-03} &
 		\textbf{5.06E-03} &
 		\textbf{1.66E-02} &
 		5.93E-02
 		\\
 		Trace: SMTL($F_1$)  vs. MTL-REG &
 		\textbf{5.06E-03} &
 		\textbf{5.06E-03} &
 		\textbf{5.06E-03} &
 		\textbf{5.06E-03} &
 		\textbf{5.06E-03}
 		\\
 		$\ell_{1,1}$: SMTL($F_1$)  vs. MTL-CLS &
 		\textbf{5.06E-03} &
 		\textbf{5.06E-03} &
 		\textbf{5.06E-03} &
 		\textbf{3.33E-01} &
 		\textbf{3.67E-02}
 		\\
 		$\ell_{1,1}$: SMTL($F_1$)  vs. MTL-REG &
 		\textbf{5.06E-03} &
 		\textbf{5.06E-03} &
 		\textbf{5.06E-03} &
 		\textbf{5.06E-03} &
 		\textbf{2.84E-02}
 		\\
 		Trace: SMTL($F_1$) vs. RAkEL &
 		\textbf{5.06E-03} &
 		\textbf{1.25E-02} &
 		\textbf{9.34E-03} &
 		9.26E-02 &
 		\textbf{5.06E-03}
 		\\
 		Trace: SMTL($F_1$) vs. MLCSSP &
 		\textbf{3.67E-02} &
 		\textbf{5.06E-03} &
 		\textbf{5.06E-03} &
 		\textbf{5.06E-03} &
 		\textbf{5.06E-03}
 		\\
 		Trace: SMTL($F_1$) vs. AdaBoostMH &
 		\textbf{5.06E-03} &
 		\textbf{5.06E-03} &
 		\textbf{5.06E-03} &
 		\textbf{5.06E-03} &
 		\textbf{5.06E-03}
 		\\
 		Trace: SMTL($F_1$) vs. HOMER &
 		\textbf{5.06E-03} &
 		\textbf{5.06E-03} &
 		\textbf{5.06E-03} &
 		\textbf{5.06E-03} &
 		\textbf{5.06E-03}
 		\\
 		Trace: SMTL(($F_1$) vs. BR &
 		\textbf{5.06E-03} &
 		\textbf{5.06E-03} &
 		\textbf{5.06E-03} &
 		1.69E-01 &
 		\textbf{5.06E-03}
 		\\
 		Trace: SMTL(($F_1$) vs. LP &
 		\textbf{5.06E-03} &
 		\textbf{5.06E-03} &
 		\textbf{5.06E-03} &
 		5.75E-01 &
 		\textbf{6.91E-03}
 		\\
 		Trace: SMTL(($F_1$) vs. ECC &
 		\textbf{5.06E-03} &
 		\textbf{5.06E-03} &
 		\textbf{5.06E-03} &
 		7.99E-01 &
 		\textbf{6.91E-03}
 		\bigstrut[b]\\
 		\hline
 	\end{tabular}%
 	
 	\label{wilcoxon_test2}%
 \end{table}%

\subsection{Results on imbalanced data}
In the scenarios of learning classifiers on imbalanced data (e.g., the number of positive training samples is much less than that of negative training samples), the metrics like F-score or AUC are more effective for evaluation than the misclassified errors. This is one of the reasons to motivate the proposed SMTL method in this paper. In MTL, the imbalance can be measured by firstly calculating the imbalance ratio in each individual task (i.e., $\frac{the\ number\ of\ positive\ instances}{the\ number\ of\ negative\ instances}$ for each task), and then averaging these ratios.

We conduct simulated experiments on 3 datasets (Segmentation, Emotions and Optdigits) to investigate the characteristics of the proposed SMTL methods on imbalanced data. In each dataset, we generate an imbalanced dataset by randomly selecting (with replacement) the positive and negative samples from the original dataset, with the ratio $1:1$, $1:5$ and $1:10$, respectively. As can be seen in Fig. \ref{ex2_AUC} and Fig. \ref{ex2_F1}, in most cases, the proposed SMTL variants consistently outperform the baseline method. For example, On Emotions with the ratio of $\frac{negative\ samples}{positive\ samples}=10:1$, the proposed SMTL indicates a relative increase of \textbf{$9.7\%$ / $12.9\%$ / $11.1\%$} over the baseline w. r. t. AUC / Macro F1 / Micro F1, respectively. In addition, with the ratio of $\frac{negative\ samples}{positive\ samples}$ increasing, the improvement of SMTL over the baseline method also increases.

\subsection{Training Time Comparison}
To investigate the training speed of the proposed method, we provide the running time comparison results in Table \ref{time}. We can see that the training time of SMTL is (less than 30 times) slower than the baseline methods. It is worth noting that the training time cost is not a critical issue in practice, because the training process is usually off-line.
%

 \begin{table}[htbp]
 	\tiny
 	\centering
 	\caption{Training Time Comparison}
	\begin{tabular}{|l|l|l|l|}
		\hline
		\multicolumn{1}{|c|}{\multirow{2}[2]{*}{\textbf{method}}} &
		\multicolumn{1}{c|}{\textbf{training time of}} &
		\multicolumn{1}{c|}{\textbf{training time of}} &
		\multicolumn{1}{c|}{\textbf{training time of}}
		\bigstrut[t]\\
		&
		\multicolumn{1}{c|}{\textbf{Optdigits}} &
		\multicolumn{1}{c|}{\textbf{Emotions}} &
		\multicolumn{1}{c|}{\textbf{Segmentation}}
		\bigstrut[b]\\
		\hline
		SMTL($\ell_{1,1}$+AUC) &
		105.200s &
		30.001s &
		1.888s
		\bigstrut[t]\\
		SMTL($\ell_{1,1}$+$F_1$) &
		510.900s &
		29.797s &
		2.964s
		\\
		MTL-CLS($\ell_{1,1}$) &
		356.200s &
		24.674s &
		2.023s
		\\
		MTL-REG($\ell_{1,1}$) &
		19.030s &
		7.427s &
		0.450s
		\\
		StructSVM &
		17.762s &
		46.468s &
		5.015s
		\\
		RAkEL &
		28.428s &
		4.117s &
		4.310s
		\\
		AdaBoostMH &
		17.157s &
		1.024s &
		0.641s
		\\
		MLCSSP &
		121.779s &
		1.563s &
		6.410s
		\\
		HOMER &
		20.643s &
		1.354s &
		0.880s
		\\
		BR &
		20.852s &
		1.859s &
		1.835s
		\\
		LP &
		16.131s &
		22.561s &
		2.103s
		\\
		ECC &
		17.852s &
		2.834s &
		1.891s
		\bigstrut[b]\\
		\hline
	\end{tabular}%

 	\label{time}%
 \end{table}%

\section{Conclusion}
In this paper, we developed Structured-MTL, a MTL method of optimizing evaluation metrics. To solve the optimization problem of Structured MTL, we developed an optimization procedure based on ADMM scheme. This optimization procedure can be applied to solving a large family of MTL problems with structured outputs.

In the future work, we plan to investigate Structured-MTL on problems other than classification (e.g., MTL for ranking problems). We also plan to improve the efficiency of Structured-MTL on large-scale learning problems.

        %





\end{document}